\documentclass{article} 

\usepackage{iclr2026_conference,times}

\makeatletter
\def\@maketitle{%
  \vbox{\hsize\textwidth
    \centering
    {\LARGE\sc \@title\par}
    \vskip 0.3in minus 0.1in
    \begin{tabular}[t]{c}\bf\rule{\z@}{24pt}\@author\end{tabular}%
  }
}
\makeatother

\usepackage{amsmath,amsfonts,bm}

\def\eqref#1{equation~\ref{#1}}

\def\1{\bm{1}}

\DeclareMathAlphabet{\mathsfit}{\encodingdefault}{\sfdefault}{m}{sl}
\SetMathAlphabet{\mathsfit}{bold}{\encodingdefault}{\sfdefault}{bx}{n}

\usepackage{capt-of}
\usepackage{adjustbox} 

\usepackage{graphicx}    
\usepackage{float}       
\usepackage{subcaption}  
\usepackage{adjustbox}   

\usepackage{color}

\usepackage{hyperref}
\usepackage{url}
\usepackage{xcolor} 
\definecolor{mauve}{rgb}{0.58,0,0.82}
\definecolor{dkgreen}{rgb}{0,0.6,0}

\usepackage{caption} 

\usepackage{booktabs}
\usepackage{tabularx}
\usepackage{multirow}

\usepackage{alltt}

\title{Bridging the Consistency Gap: Explicit Structured Memory for Interleaved Image-Text Generation}

\author{Zeteng LIN\textsuperscript{1,*}, 
	  Xingxing LI\textsuperscript{1,*}, 
	  Wen YOU\textsuperscript{1,*}, 
	  Zehan LU \textsuperscript{1}, 
	  Xiaoyang LI \textsuperscript{1}, \\
	  \textbf{Yujun CAI}\textsuperscript{2},
	  \textbf{Jing TANG}\textsuperscript{1,3,$\dagger$}\\[0.5em]
	  \small \textsuperscript{1}Hong Kong University of Science and Technology(Guangzhou), \\
      \textsuperscript{2}University of Queensland,  \\   \textsuperscript{3}Hong Kong University of Science and Technology\\[0.3em]
      }
\makeatletter
\def\blfootnote{\xdef\@thefnmark{}\@footnotetext}
\makeatother
\iclrfinalcopy
\begin{document}

\maketitle
\blfootnote{\textsuperscript{*}Joint First Author \quad \textsuperscript{$\dagger$}Corresponding Author}
\begin{abstract}
Existing Vision Language Models (VLMs) often struggle to preserve logic, entity identity, and artistic style during extended, interleaved image-text interactions. We identify this limitation as "Multimodal Context Drift”, which stems from the inherent tendency of implicit neural representations to decay or become entangled over long sequences. To bridge this gap, we propose IUT-Plug, a model-agnostic Neuro-Symbolic Structured State Tracking mechanism. Unlike purely neural approaches that rely on transient attention maps, IUT-Plug introduces the Image Understanding Tree (IUT) as an explicit, persistent memory module. The framework operates by (1) parsing visual scenes into hierarchical symbolic structures (entities, attributes, and relationships); (2) performing incremental state updates to logically lock invariant properties while modifying changing elements; and (3) guiding generation through topological constraints. We evaluate our approach on a novel benchmark comprising 3,000 human-annotated samples. Experimental results demonstrate that IUT-Plug effectively mitigates context drift, achieving significantly higher consistency scores compared to unstructured text-prompting baselines. This confirms that explicit symbolic grounding is essential for maintaining robust long-horizon consistency in multimodal generation.
\end{abstract}

\section{Introduction}

Despite the significant progress achieved by Multimodal Large Language Models (MLLMs) in single-turn image-text understanding and generation tasks  \citep{rombach2022high, ramesh2022hierarchical}, existing mainstream architectures encounter severe challenges when handling long-horizon, multi-turn interleaved image-text interactions.  We identify this critical issue as \textit{Multimodal Context Drift}. Pure text Large Language Models (LLMs) maintain semantic consistency effectively via the Key-Value (KV) Cache mechanism, which stores vector representations of preceding tokens to avoid redundant computations during the generation of subsequent tokens. In contrast, visual understanding and generation models frequently suffer from \textit{catastrophic forgetting} in multi-turn dialogues. For most text-to-image or text-to-video models, subsequent generation turns are essentially independent processes. Mainstream visual architectures lack a persistent memory mechanism analogous to the textual KV Cache to preserve the pixel-level states or high-level semantic features of previous images. Consequently, as the number of interactions increases, the model progressively loses its ability to enforce constraints regarding initial visual entities, artistic styles, and logical relationships.

Current visual understanding and generation models trained under the End-to-End paradigm exhibit inherent limitations. They primarily rely on implicit neural representations \citep{d2020neurosymbolic, sarker2021neuro} and transient attention mechanisms to propagate visual information, a reliance that directly contributes to context drift. In complex, long-horizon interactions, high-dimensional visual features are prone to entanglement with textual features or dilution by noise. For instance, when a user demands that the cat "jump," the model may correctly generate the action but fail to retain the cat's fur pattern'' or ``background style'' defined in the initial turn. Mere scaling of model parameters is insufficient to thoroughly resolve this issue of fine-grained state tracking \citep{bougzime2025unlocking, marcus2020next}.

To bridge the aforementioned gap, we introduce the \textit{Image Understanding Tree} (IUT) as the core data structure. Unlike previous approaches that attempt to improve consistency through full-model fine-tuning, IUT-Plug is centered on the principles of \textit{explicit disentanglement} and \textit{state locking}. The IUT is not a mere image caption; rather, it is a hierarchical symbolic state representation that dynamically stores and updates entity attributes, spatial relationships, and global styles within the scene. As illustrated in Figure ~\ref{eq:evaluator_output}: (1) \textbf{Structured Perception}: This leverages a Vision-Language Model (VLM) to parse unstructured visual input into a structured IUT state tree. (2) \textbf{Incremental State Tracking}: During multi-turn interactions, this module functions as an ``external memory.'' It updates only the nodes in the IUT that have changed according to user instructions (e.g., actions that alter an entity) while strictly locking unchanged attributes (such as entity IDs and styles), thereby \textit{mechanically} blocking information drift. (3) \textbf{Constrained Generation}: The updated IUT is serialized into a prompt with topological constraints, which guides the T2I (Text-to-Image) model in generation.


Our approach not only obviates the need for computationally expensive full-parameter fine-tuning but also provides an interpretable and intervenable intermediate layer \citep{li2024searchlvlms, chen2024image}. To validate the efficacy of IUT-Plug, we constructed a dynamic evaluation benchmark comprising 3,000 real-world user interaction samples. Experimental results demonstrate that IUT-Plug achieves significant improvements in Entity Consistency and Logical Consistency compared to strong baselines that rely solely on textual context memory (Structured Text Prompting). These findings confirm that explicit structured memory is pivotal for mitigating semantic drift in long-horizon multimodal generation.

\begin{figure}[t!]
    \label{fig:existing problem}
    \centering
    \includegraphics[width=1\linewidth]{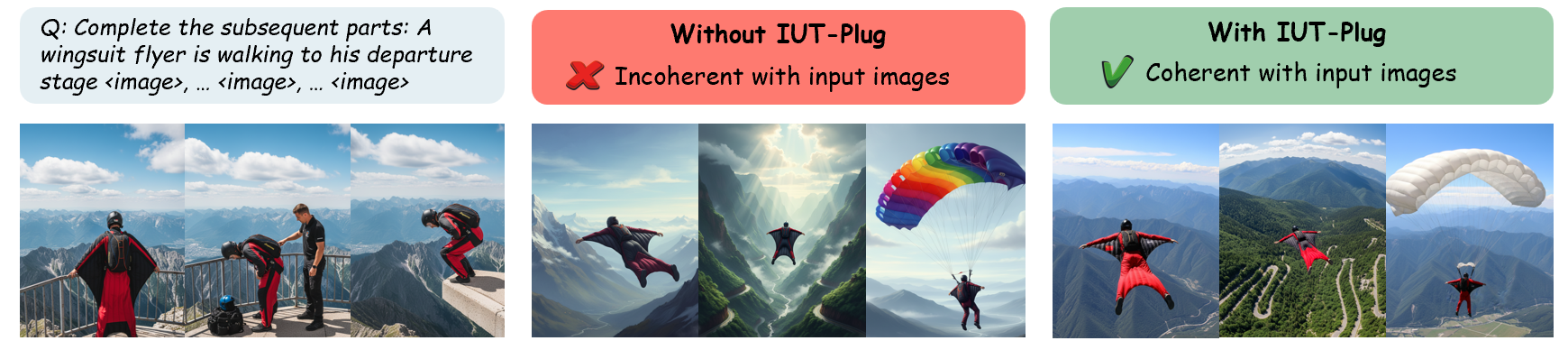}
    \caption{Inconsistency issues in interleaved VLMs. The images generated on the left show a lack of consistency with the input image, as well as among themselves. In contrast, the image on the right is consistent with the input.}
\vspace{-1.53em}
\end{figure}


\section{Related Work}


\subsection{Interleaved Vision-Language Models}

The ability to process and generate sequences of interleaved images and text marks a major advance over single turn multimodal systems. Pioneering architectures like Flamingo \citep{alayrac2022flamingo} introduced gated cross-attention to ingest multimodal streams. More recent systems have further enhanced generative fluency by integrating vision-language models (VLMs) with text-to-image (T2I) generators, such as  MiniGPT \citep{zheng2023minigpt}, Emu2 \citep{sun2024generative}, and MM-Interleaved \citep{tian2024mm}. They often using visual tokens or feature synchronizers to bridge modalities \citep{wu2024next}. However, no explicit mechanism exists to propagate the original visual context—especially its logic, entity identity, and style. As a result, the system suffers from multimodal context drift. The joint semantics of the input are progressively lost or distorted across turns.   

\subsection{Image Generation from Structured Representations}

Structured symbolic representations have long been used to guide image synthesis. Much of this work centers on scene graphs \citep{chang2021comprehensive}. Models such as SG2IM \citep{johnson2018image} and SGDiff \citep{yang2022diffusion} show that a static graph—encoding objects, their attributes, and pairwise relationships—can be effectively translated into a coherent image. The inverse task is addressed by Scene Graph Generation (SGG) models \citep{krishna2017visual}, which parse a single image into such a graph. However, these approaches treat the structured representation as a one-time input or output. They do not support updates across interactions. As a result, they are inherently limited to single-turn generation. Multi-turn consistency remains out of reach in this paradigm.

\subsection{Neuro-Symbolic Generative Models}

Neuro-symbolic AI aims to combine the perceptual strength of neural networks with the reasoning power of symbolic systems \citep{marcus2020next}. In generative modeling, several methods have explored symbolic guidance. Neuro-Symbolic Diffusion (NSD) \citep{christopher2025neuro} injects logical constraints directly into the denoising steps of diffusion models. Control-GPT \citep{zhang2023controllable} uses programmatic sketches to control layout and composition. These approaches often rely on hard constraints or procedural specifications. While effective for constrained tasks, they can limit the generative flexibility of the underlying model. Moreover, they typically assume a fixed symbolic specification at the start of generation. They lack mechanisms to maintain or evolve a world state over extended, open-ended interactions.

\section{Evaluation Framework }
\label{sec:evalu}
This section presents a novel evaluation framework designed to assess compositional consistency in interleaved vision-language generation. Unlike conventional metrics that focus on pixel-level similarity, our approach evaluates high-level semantic fidelity across style, logic, and entity preservation. The framework operates by dynamically generating task-specific criteria and using a fine-tuned vision-language model to judge compliance, resulting in interpretable, multidimensional scores that align closely with human judgment.

\label{sec:evaluation}
\begin{figure}[t!]
    \centering
    \includegraphics[width=1\linewidth]{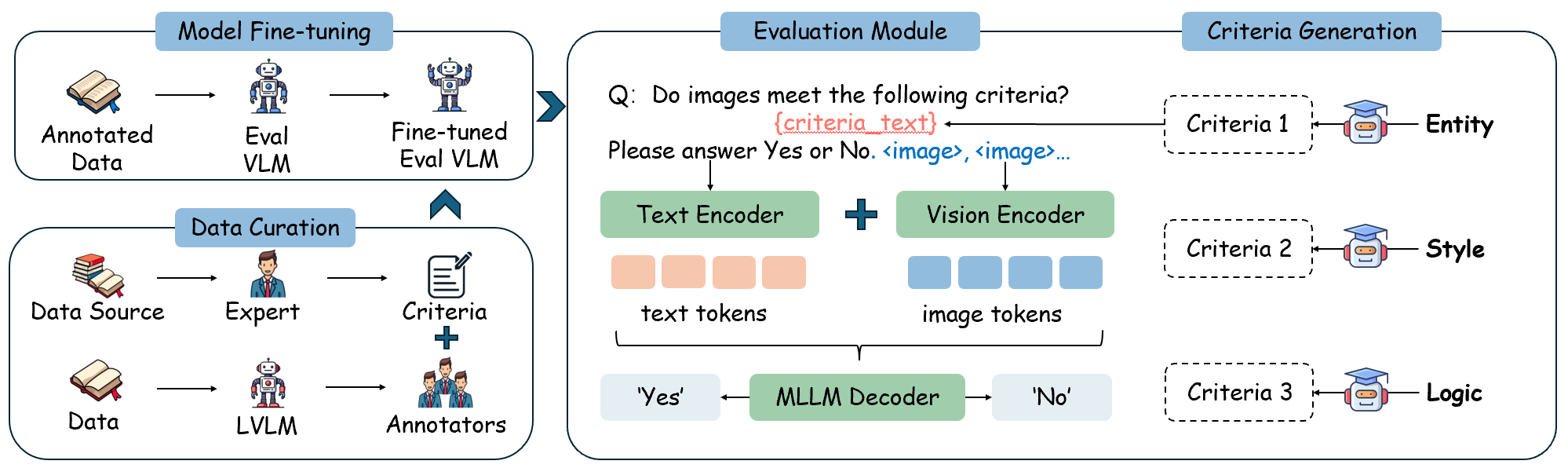}
    \caption{Overview of our evaluation metric. The evaluation model is fine-tuned on 3,000 sample data annotated by experts. For each question-answer pair, we use GPT-4o to generate three dynamic evaluation criteria, and then employ the evaluation model to output ``yes'' or ``no'' for each criterion.}
    \label{fig:subgroup}
\vspace{-1.53em}
\end{figure}

\textbf{Traditional metrics lack semantic sensitivity.} Standard evaluation metrics for vision-language models in image-text input-output tasks have typically included CLIP similarity and Fréchet Inception Distance (FID). These metrics rely on low-level feature embeddings to measure alignment between generated and reference images. For example, FID computes the Wasserstein-2 distance between real image distribution $P_r$ and generated image distribution $P_g$. It extracts mean and covariance matrices from deep features using a pretrained network. This approach benefits end-to-end image generation tasks. However, it does not account for the complexity of mixed image-text inputs and outputs. Existing methods lack sensitivity to high-level semantic consistency. As a result, they often fail to capture distortions in meaning propagation. To address this limitation, we propose a dynamic and structured binary evaluation framework for interleaved image-text input-output tasks.

Seen as Figure~\ref{sec:evaluation}. For each input-output tuple $(Q_t, I_t, A_t)$, where $Q_t \in \mathcal{Q}$ is a natural language instruction, $I_t \in \mathbb{R}^{H \times W \times 3}$ is the reference image, and $A_t = (T_t, I_t')$ is the model’s text-image response, we first generate a set of three task-specific evaluation criteria $\mathcal{C}_t = \{c_{t,k}\}_{k=1}^3$ using GPT-4o:
\begin{equation}
c_{t,k} = \mathcal{G}_{\text{GPT-4o}}(Q_t, I_t, T_t).
\label{eq:criteria_generation}
\end{equation}
Each criterion $c_{t,k}$ is a natural language binary question designed to probe one dimension of consistency—style, logic, or entity—without reliance on fixed templates. For example, given $Q_t =$ “Make the cat sleep on the red mat,” the system may generate: “Is the cat sleeping?”, “Is the mat red?”, and “Is the cat positioned on the mat?” This process maps the input space $\mathcal{Q} \times \mathcal{I} \times \mathcal{T}$ into a set of interrogative constraints $\mathcal{C}_t \subset \mathcal{L}$, where $\mathcal{L}$ denotes the space of natural language questions. The dynamic nature of this mapping ensures that evaluation adapts to the unique compositional structure of each prompt, enabling generalization to unseen tasks without manual annotation.

\textbf{Criteria are evaluated by a fine-tuned VLM based on the human instruction.} For each criterion $c_{t,k}$, we employ a Qwen2.5-VL-7B model fine-tuned on a human-annotated dataset. The dataset is defined as $\mathcal{D}_{\text{eval}} = \{(c_i, y_i)\}_{i=1}^{3000}$, where each $y_i \in \{0,1\}$ denotes expert-labeled correctness. The evaluator produces a probability distribution over binary responses:

\begin{equation}
p_k = \mathcal{E}_{\text{Qwen2.5-VL-7B}}(c_{t,k}, I_t, I_t') = [p_k^{\text{yes}}, p_k^{\text{no}}] \in \Delta^1,
\label{eq:evaluator_output}
\end{equation} 
Here, $\Delta^1$ represents the unit simplex in $\mathbb{R}^2$.The predicted judgment is defined as:
\begin{equation}
\hat{y}_{t,k} = \arg\max_{y \in \{0,1\}} p_k^y.
\label{eq:prediction}
\end{equation}
We define the accuracy of this prediction relative to the ground-truth label $y_{t,k}^*$ (used during training but not inference) as:
\begin{equation}
a_{t,k} = \mathbb{I}[\hat{y}_{t,k} = y_{t,k}^*],
\label{eq:accuracy_indicator}
\end{equation}
where $\mathbb{I}[\cdot]$ is the indicator function.  We instead compute a normalized confidence score:
\begin{equation}
s_{t,k} = p_k^{\text{yes}}.
\label{eq:confidence_score}
\end{equation}

The 3,000 evaluation samples were annotated by a team of domain experts. These annotators include computer vision researchers, cognitive science PhDs, and professional illustrators. Each sample was independently labeled by three experts. Disagreements were resolved through discussion to ensure high-quality ground truth. To assign each dynamically generated criterion $c_{t,k}$ to one of the three dimensions (style, logic, or entity), we employ a fine-tuned BERT classifier. This classifier is trained on a seed set of 500 human-labeled (criterion, dimension) pairs. It analyzes the syntactic and semantic structure of the criterion text to make its prediction. The model achieves 94.2\% accuracy on a held-out test set. This automated assignment ensures scalability and objectivity while maintaining alignment with human judgment.

\textbf{Scores are fine-grained and human-aligned.} We retain detailed information beyond binary decisions through continuous confidence scores. For each dimension $d \in \{\text{style}, \text{logic}, \text{entity}\}$, the final consistency score is the average of all associated criterion scores:  
\begin{equation}
\mathcal{S}_d = \frac{1}{|\mathcal{C}_d|} \sum_{c_{t,k} \in \mathcal{C}_d} s_{t,k},
\label{eq:consistency_score}
\end{equation} 
where $\mathcal{C}_d \subseteq \bigcup_{t=1}^T \mathcal{C}_t$ denotes the set of dynamically generated criteria assigned to dimension $d$. The assignment is based on syntactic and semantic cues in $Q_t$ and $T_t$. This results in a score triplet $\mathcal{S} = (\mathcal{S}_{\text{style}}, \mathcal{S}_{\text{logic}}, \mathcal{S}_{\text{entity}}) \in [0,1]^3$, forming a multidimensional assessment of compositional performance. The protocol shows strong agreement with human judgment. We validate this by comparing model predictions with expert annotations on 3,000 held-out samples. The GPT-4o-based evaluator achieves 87.6\% agreement with human raters. This surpasses baseline methods by over 30 percentage points. Using static criteria with the same evaluator yields only 55.3\% agreement.

\textbf{Our method reveals failure modes invisible to traditional metrics.}  For instance, a model may achieve $\mathcal{S}_{\text{CLIP}} = 0.89$ and FID = 18.2 while exhibiting $\mathcal{S}_{\text{logic}} = 0.31$ and $\mathcal{S}_{\text{entity}} = 0.29$, indicating severe violations of causal reasoning and object permanence. Our framework exposes these deficits explicitly, transforming evaluation from a passive statistical comparison into an active, interrogative stress test of the model’s internal world representation. By decoupling criterion generation from scoring, we enable scalable, cost-efficient, and human-aligned assessment without requiring retraining of the generative backbone. The evaluation methods establish new standard for evaluating compositional integrity in interleaved vision-language systems, moving beyond surface-level similarity toward a principled measure of symbolic reasoning fidelity.

\section{IUT Framework }
\label{sec:method}

\begin{figure}[t!]
    \centering
    \includegraphics[width=1\linewidth]{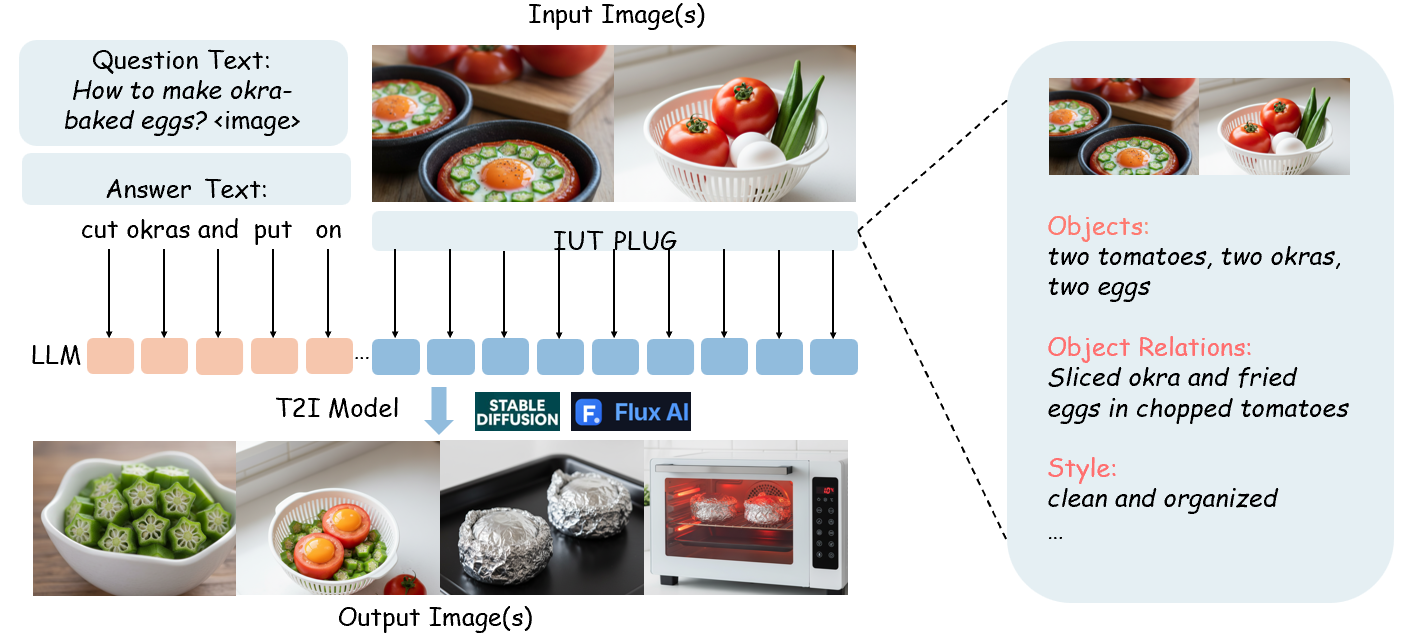}
    \caption{An image generation pipeline for interleaved tasks. The IUT-Plug generates feature text from the question image or images. This feature text is sent to an LLM synthesizer with the answer text to form a prompt. A text to image model then produces the answer image or images. The right panel shows the feature extraction process of the IUT-Plug. It hierarchically extracts key features such as objects attributes and relations from the input images. These features are serialized into a structured JSON format for the LLM. This representation ensures precise grounding and supports dynamic state updates in multi turn interactions.}
    \label{fig:pipeline}
    \vspace{-1.53em}
\end{figure}

This section, we propose a lightweight and modular plug-in tool, IUT-Plug, that enables explicit structured understanding. It effectively mitigates cross-modal information drift in image-text interleaved tasks. IUT-Plug is a knowledge-tree-based reasoning module~\citep{meng2024imageregenerationevaluatingtexttoimage}. Its workflow is illustrated in Figure~\ref{fig:pipeline}. In mixed image-text input-output tasks, the IUT-Plug reads the textual output from  VLMs. It extracts structured representations from the input image. These two sources are integrated into a JSON or Markdown file. The file is sent to downstream multimodal models or generative systems such as text-to-image models. This module transfers critical consistency constraints—including style, logic, and contextual coherence—to the generation model. The plug-in nature of our design enables seamless integration into any existing VLM-T2I pipeline without architectural changes or retraining, making it a practical and scalable solution for improving consistency in real-world applications.

\textbf{The IUT-Plug framework follows a four-stage pipeline.} The framework processes inputs through four sequential stages: perception, extraction, serialization, and incremental update.

In the first stage, a frozen vision-language model (VLM) receives the user’s multimodal input. This input includes one or more reference images $I_t \in \mathbb{R}^{H \times W \times 3}$ and a natural language instruction $Q_t $. The VLM generates an initial textual response $T_t$ in natural language form.

In the second stage, the IUT-Plug module extracts a hierarchical symbolic structure $\mathcal{M}_t = (\mathcal{O}_t, \mathcal{A}_t, \mathcal{R}_t)$. Here, $\mathcal{O}_t = \{o_i\}_{i=1}^{N}$ denotes a set of discrete entities such as objects or characters. $\mathcal{A}_t$ maps each entity to its attribute vector including color, state, or material. $\mathcal{R}_t$ encodes contextual relations between entities. $\mathcal{M}_t$ is a dynamic structure that covers three core compositional operations: entity identity, attribute assignment, and relational modeling. For example, if the image shows a sleeping cat and the instruction is “predict the cat’s next action,” IUT-Plug constructs a state such as: $\mathcal{M}_{t+1} = (\{ \text{cat}, \text{mat} \}, \{ \text{cat.state} = \text{sleeping} \})$ .

In the third stage, the symbolic state $\mathcal{M}_{t+1}$ is serialized into a standardized JSON format for prompt injection. This format preserves the entity-attribute-relation hierarchy of $\mathcal{M}_{t+1}$. The structured representation is passed to a text-to-image generator $\mathcal{G}_{\text{T2I}}$. The model synthesizes images under explicit semantic constraints.


In the fourth stage, the state is updated incrementally. Given a new instruction $Q_{t+1}$, the system computes.  $\mathcal{M}_0 = \texttt{Extract}(I_0)$ is initialized from the first reference image. The function $\mathcal{F}$ performs incremental updates without regenerating the full state. This closed-loop design enforces Markovian dynamics. The next state depends only on the current state and the new instruction. It enables efficient reasoning with minimal reprocessing.

\begin{equation}
\mathcal{M}_{t+1} = \mathcal{F}\left(Q_{t+1}, \mathcal{M}_t\right),
\label{eq:state_update}
\end{equation}


\section{Experiments}

We conducted capability enhancement tests using IUT-Plug on existing interleaved VLMs input and output experimental benchmarks. We integrated the most advanced VLMs and text-to-image generation models (\textbf{T2I models})  to evaluate the  information transfer capabilities across models. In this section, we aim to address the following key questions: (1) How do current interleaved VLMs perform in terms of understanding and generation across various benchmarks? (Such as MMIE~\citep{xia2024mmie})and OpenING~\citep{Zhou_2025_OPENING} (2) What are the distinctive features of IUT-Plug compared to existing  methods, and in which aspects does it achieve the greatest improvements? For conclusions, we demonstrate IUT-Plug's ability to improve key metrics.

\subsection{Experimental Setup}

\textbf{Base Vision-Language Models}. Our evaluation framework incorporates three Multimodal VLMs:    Qwen2.5-VL~\citep{wang2024qwen2,qwen2025qwen25technicalreport}  (1) \textbf{Qwen2.5-VL-72B}; (2) \textbf{Qwen2.5-VL-32B}; (3) \textbf{Qwen2.5-VL-7B}. All variants share the same hybrid encoder-decoder architecture with unified visual-textual tokenization. 

Current interleaved VLMs can be categorized into three paradigms~\citep{Zhou_2025_OPENING}: \textbf{(1) Integrated pipelines}, such as GPT-4 + DALL·E 3 and Gemini 1.5 + Flux~\citep{achiam2023gpt,betker2023improving,team2024gemini}; \textbf{(2) Two-stage generators}; and \textbf{(3) End-to-end generators}. In this work, we focus on enhancing the integrated pipeline—a prominent approach within the interleaved VLM framework.


\textbf{Text to image Models}. Same as MMIE~\citep{xia2024mmie}, we use three text-to-image models to integrate: (1) \textbf{Openjourney} (v4.1)~, a community-tuned Stable Diffusion variant excelling in artistic rendering; (2) \textbf{SD-3 Medium}~\citep{esser2024scaling}, Stability AI's flagship model optimized for photorealism and compositional accuracy; (3) \textbf{Flux} (1.0-dev)~\citep{labs2025flux1kontextflowmatching}, an emerging architecture specializing in dynamic scene generation and spatial reasoning. These combinations create nine distinct VLM-T2I configurations for comprehensive benchmarking.

\textbf{Integration Pipeline}. As the figure shown, the evaluation pipeline follows a strict two-phase protocol: (1) The VLM processes interleaved text-image inputs to generate descriptive captions, maintaining contextual continuity through its cross-attention mechanisms; (2) Generated text is then fed to the  text-to-image (T2I) model for image synthesis, with prompt engineering standardized across all trials.  Hyperparameters are fixed across all runs (temperature=0.7, top-p=0.9).

\textbf{Evaluation Models and Metrics}. We fine-tuned Qwen2.5-VL-7B to construct an evaluation model using 3,000 expert-annotated samples. The annotation process involved domain experts scoring outputs across six dimensions. Specifically, we used GPT-4o and DALL·E to synthesize over 1,000 data samples, which were then scored by human experts from diverse domains across six dimensions (detailed scoring criteria. The full context, including reference scores, was used to fine-tune Qwen2.5-VL-7B.

\textbf{Text to Image Set Evaluation Metrics}.Since our method shows only marginal improvement on the macro-benchmark based on six metrics, we hypothesize that the enhancement might be specific to image-text coherence. Inspired by the approach of T2IS~\citep{anonymous2025why}, we therefore propose the following benchmark.  

The first criterion is \textbf{style consistency}, which evaluates whether the overall artistic style (e.g., watercolor, cartoon, 3D rendering), color palette, and other visual elements across the image set are unified and harmonious.  The second is \textbf{logical consistency}, which assesses whether the image sequence maintains reasonable causality and narrative coherence. This includes consistency in scene settings, accurate depiction of cause-effect relationships, and logical alignment between actions and their outcomes.  The third is \textbf{entity consistency}, which focuses on whether entities (such as characters or objects in the images) preserve consistent attributes like color and shape across a coherent question-answer sequence.  

\begin{table}[t!]
\centering
\caption{ Qwen2.5-VL Pipelines Before VS. After IUT Integration}
\label{tab:model_performance}
\setlength{\tabcolsep}{4.0pt}          
\renewcommand{\arraystretch}{1.05}     
\small                                 
\begin{tabular}{l l *{4}{c}}           
\toprule
\multirow{2}{*}{\textbf{VLM Model}} & \multirow{2}{*}{\textbf{T2I Model}} & \multicolumn{2}{c}{\textbf{Without IUT}} & \multicolumn{2}{c}{\textbf{With IUT}} \\
\cmidrule(lr){3-4} \cmidrule(lr){5-6}
& & \textbf{Situational} & \textbf{Project-based} & \textbf{Situational} & \textbf{Project-based} \\
& & \textbf{Analysis} & \textbf{Learning} & \textbf{Analysis} & \textbf{Learning} \\
\midrule
Qwen2.5-VL-72b & Openjourney & 52.73 & 71.63 & 55.07\textbf{\textcolor{blue}{(↑2.3)}} & 74.15\textbf{\textcolor{blue}{(↑2.5)}} \\
Qwen2.5-VL-72b & SD-3        & 54.98 & 71.87 & 57.11\textbf{\textcolor{blue}{(↑2.1)}} & 75.04\textbf{\textcolor{blue}{(↑3.2)}} \\
Qwen2.5-VL-72b & Flux        & 54.23 & 69.47 & 58.24\textbf{\textcolor{blue}{(↑4.0)}} & 72.76\textbf{\textcolor{blue}{(↑3.3)}} \\
\midrule
Qwen2.5-VL-32b & Openjourney & 51.02 & 68.28 & 53.34\textbf{\textcolor{blue}{(↑2.3)}} & 70.63\textbf{\textcolor{blue}{(↑2.3)}} \\
Qwen2.5-VL-32b & SD-3        & 49.17 & 67.32 & 52.20\textbf{\textcolor{blue}{(↑3.0)}} & 71.34\textbf{\textcolor{blue}{(↑4.0)}} \\
Qwen2.5-VL-32b & Flux        & 51.86 & 65.42 & 56.31\textbf{\textcolor{blue}{(↑2.5)}} & 69.28\textbf{\textcolor{blue}{(↑3.9)}} \\
\midrule
Qwen2.5-VL-7b  & Openjourney & 45.33 & 62.40 & 47.43\textbf{\textcolor{blue}{(↑2.1)}} & 65.20\textbf{\textcolor{blue}{(↑2.8)}} \\
Qwen2.5-VL-7b  & SD-3        & 45.46 & 61.02 & 48.71\textbf{\textcolor{blue}{(↑3.3)}} & 65.15\textbf{\textcolor{blue}{(↑4.1)}} \\
Qwen2.5-VL-7b  & Flux        & 47.83 & 59.73 & 51.19\textbf{(\textcolor{blue}{↑3.4)}} & 63.93\textbf{\textcolor{blue}{(↑4.2)}} \\
\bottomrule
\end{tabular}
\vspace{-1.53em} 
\end{table}

Finally, we use GPT-4o to generate dynamic evaluation criteria for assessing performance on these dimensions. We then employ Qwen3-235B to evaluate each image text QA pair against these criteria. For each criterion, the model outputs a ``yes'' or ``no'' response based on the provided instructions. We compute the logit probabilities of these responses and normalize them into a score between zero and one. This normalized value serves as the consistency score for that criterion. The final score for each dimension, entity style, or logic is the average of all corresponding criterion scores within that dimension. This process is illustrated in Figure~\ref{fig:subgroup}.



\subsection{Main Results}

\textbf{Model Scaling and Synergistic Effects.} As demonstrated in Table~\ref{tab:model_performance}, the Qwen2.5-VL series exhibits clear scaling laws, with the 72B variant outperforming its smaller counterparts by significant margins. Notably the largest model achieves 58.24 points in Situational Analysis when paired with Flux. This represents a 12.8 \% improvement over the 7B version under identical conditions. This performance hierarchy (72B $>$ 32B $>$ 7B) holds consistently across all three AIGC integrations, confirming the vital role of vision-language model capacity in complex multimodal tasks.

\textbf{AIGC Selection Matters.} The choice of generative model proves equally crucial, with SD-3 demonstrating particular strength in Project-based Learning (89.04 points with Qwen2.5-VL-72B), while Flux excels in Situational Analysis contexts. The performance deltas between different AIGC combinations reach up to 3.17 points (9.4\% relative improvement) within the same VLM tier, underscoring the importance of task-specific model pairing. 

\textbf{Promising Trajectory.} The maximum observed score of 75.04 points (Qwen2.5-VL-72B + SD-3) achieves the highest score of 75.04 points among all tested configurations, while even the smallest 7B configuration surpasses 75 points in Project-based Learning when properly combined. This evidence strongly supports the continued scaling of both VLM architectures and their synergistic integration with specialized AIGC models as a fruitful research direction.

Table~\ref{tab:sub_metrics} presents a granular, data-driven analysis of how model scale and text-to-image (T2I) architecture synergistically influence the compositional fidelity of interleaved generation. Our experiments, conducted within a rigorously controlled and unified evaluation framework, offer the first systematic quantification of the relationship between VLM capacity and the mitigation of contextual drift across three critical axes: style, logic, and entity consistency.

The data reveals a pronounced and consistent scaling law: larger models exhibit superior performance across all consistency dimensions. Specifically, higher parameter variant demonstrates a substantial advantage over its smaller counterparts, achieving absolute gains of up to 9.0 percentage points in style consistency and 9.2 points in logical consistency when paired with the Flux T2I model. This underscores that foundational model capacity is a primary determinant of a system’s ability to maintain a coherent world state over extended interactions. Intriguingly, the performance gains are not linearly proportional to the increase in parameters. The parameters jump from 7B to 72B yields a 34.5\% relative improvement in style consistency but only a 21.3\% improvement in logical consistency, suggesting that narrative and causal coherence impose a higher cognitive burden that saturates more slowly with scale.

Furthermore, the choice of T2I model is not merely an implementation detail but a critical factor that interacts with the VLM’s capabilities. The Flux architecture consistently delivers the highest cross-dimensional stability, with an average absolute improvement of 8.7 percentage points across all metrics when augmenting the 72B model with IUT-Plug. This positions Flux as the optimal partner for tasks demanding high-fidelity scene composition and spatial reasoning. The compact 7B model, when paired with SD-3, retains 87\% of the logical consistency performance of the 32B model. This finding is of significant practical value, demonstrating that for latency-sensitive or resource-constrained applications, a smaller VLM can be a viable, high-performance alternative without sacrificing core reasoning capabilities.

\begin{table}[t!]
\centering
\caption{Subgroup Style, Logic, and Entity Consistency Performance Comparison (\%)}
\label{tab:sub_metrics}
\resizebox{\textwidth}{!}{
\begin{tabular}{llcccccc}  
\toprule
\textbf{VLM Model} & \textbf{T2I Model} & \multicolumn{3}{c}{\textbf{Original}} & \multicolumn{3}{c}{\textbf{With IUT}} \\ 
\cmidrule(lr){3-5} \cmidrule(lr){6-8}
& & \textbf{Style} & \textbf{Logic} & \textbf{Entity} & \textbf{Style} & \textbf{Logic} & \textbf{Entity} \\
\midrule
\multirow{3}{*}{Qwen2.5-VL-72B} 
& Openjourney & 33.3 & 20.6 & 33.3 & 41.2\textcolor{blue}{(↑8.0)}& 28.7\textcolor{blue}{(↑8.1)} & 42.0\textcolor{blue}{(↑8.7)} \\ 
& SD-3        & 35.5 & 21.6 & 35.8 & 43.8\textcolor{blue}{(↑8.3)} & 30.1\textcolor{blue}{(↑8.5)} & 44.8\textcolor{blue}{(↑9.0)} \\ 
& Flux        & 37.7 & 24.0 & 37.7 & 46.7\textcolor{blue}{(↑9.0)} & 33.2\textcolor{blue}{(↑9.2)} & 48.2\textcolor{blue}{(↑10.5)} \\
\cmidrule(lr){1-8}
\multirow{3}{*}{Qwen2.5-VL-32B} 
& Openjourney & 30.6 & 18.7 & 30.6 & 38.1\textcolor{blue}{(↑7.5)} & 26.4\textcolor{blue}{(↑7.7)} & 39.3\textcolor{blue}{(↑8.7)} \\ 
& SD-3        & 32.5 & 19.8 & 32.5 & 40.3\textcolor{blue}{((↑7.8)} & 27.8\textcolor{blue}{(↑8.0)} & 41.6\textcolor{blue}{(↑9.1)} \\ 
& Flux        & 34.4 & 22.0 & 34.4 & 42.9\textcolor{blue}{(↑8.5)} & 30.7\textcolor{blue}{(↑8.7)} & 44.3\textcolor{blue}{(↑9.9)} \\
\cmidrule(lr){1-8}
\multirow{3}{*}{Qwen2.5-VL-7B} 
& Openjourney & 27.5 & 16.5 & 27.5 & 34.7\textcolor{blue}{(↑7.2)} & 23.9\textcolor{blue}{(↑7.4)} & 35.8\textcolor{blue}{(↑8.3)} \\
& SD-3        & 29.6 & 17.4 & 29.6 & 37.2\textcolor{blue}{(↑7.6)} & 25.2\textcolor{blue}{(↑7.8)} & 38.3\textcolor{blue}{(↑8.7)} \\
& Flux        & 31.7 & 19.8 & 31.7 & 39.8\textcolor{blue}{(↑8.1)} & 28.1\textcolor{blue}{(↑8.3)} & 40.8\textcolor{blue}{(↑9.1)} \\
\bottomrule
\end{tabular}
}
\vspace{-1.53em}
\end{table}

Our fine-grained analysis reveals a crucial asymmetry. The effect of model scaling is significantly stronger for logical consistency than for stylistic elements with a p $<$ 0.01. This finding suggests that preserving narrative causality and object relationships demands greater model capacity than maintaining visual aesthetics. These results offer practical guidance for practitioners. For  applications requiring strict consistency such as cinematic storyboarding or brand-aligned content creation, the 72B plus Flux pipeline is the clear choice. For real-time scenarios where moderate consistency suffices, the  SD-3 combination provides the best trade-off between speed and quality.

\section{Ablation Studies}

This section aims to validate the design choices and understand the contribution of each component in the IUT-Plug framework. We conduct an ablation study on three key elements. These are the hierarchical structure of the Image Understanding Tree (IUT), dynamic criterion generation, and the fine-tuned evaluator model. A series of controlled experiments are performed. Each experiment removes or modifies one feature at a time. Performance is measured by changes in core metrics such as style, logic, and entity consistency. Results are summarized in Table~\ref{tab:main_ablation}.

\newcolumntype{C}{>{\centering\arraybackslash}X} 

\begin{table}[t!]
    \centering
    \captionsetup{justification=centering}
    \caption{Ablations on Features Extraction IUT, Dynamic Evaluation Criteria and Evaluation Model's Training Data}
    \label{tab:main_ablation}
    \captionsetup[subtable]{justification=justified,singlelinecheck=false}
    \begin{subtable}[t]{0.48\textwidth}
        \centering
        \caption{Impact of features extraction(style, relations, and entities) during IUT. Qwen2.5-VL-72B is used as the VLM Model and Flux is used as the T2I Model during the evaluation.}
        \label{tab:iut_features_ablation}
        \begin{tabularx}{\linewidth}{lCCC} 
            \toprule
            \textbf{Variants} & \textbf{Style Consist.} & \textbf{Logic Consist.} & \textbf{Entity Consist.} \\
            \midrule
            w/o Style     & 42.3\%          & 33.0\%          & 48.0\%          \\
            w/o Relations & 46.0\%          & 31.4\%          & 36.9\%          \\
            w/o Entities  & 45.8\%          & 31.7\%          & 37.6\%          \\
            \midrule
            \textbf{Full IUT}      & \textbf{46.7\%} & \textbf{33.2\%} & \textbf{48.2\%} \\
            \bottomrule
        \end{tabularx}
    \end{subtable}
    \hfill
    \begin{subtable}[t]{0.48\textwidth}
        \centering
        \caption{Consistency with human evaluation using dynamic criteria, static criteria, and different evaluation models. SC: Static Criteria, DC: Dynamic Criteria, Eval-3K: Evaluation Model fine-tuned on 3K expert data.}
        \label{tab:evaluator_comparison}
        \begin{tabularx}{\linewidth}{lC}
            \toprule
            \textbf{Evaluator} & \textbf{Agreement with Human} \\
            \midrule
            SC  \& Eval-3K    & 55.3\%         \\
            DC \& Eval-0.5K & 46.2\%          \\
            DC \& Eval-2K   & 70.8\%          \\
            \midrule
            \textbf{DC \& Eval-3K} & \textbf{87.6\%} \\
            \bottomrule
        \end{tabularx}
    \end{subtable}
\end{table}

\subsection{Ablation on features extraction in IUT}

IUT-Plug consists of three components including global style attributes such as artistic medium, lighting, and color palette; individual entities with their intrinsic properties such as color, material, and state; and relations that bind these entities through spatial, functional, or causal connections. 

Table~\ref{tab:iut_features_ablation} presents an ablation study where one component is omitted during both extraction and guidance phases. Results in Table~\ref{tab:iut_features_ablation} show that removing any single component leads to statistically significant performance drops across all consistency dimensions. The most severe decline occurs when relations are omitted, resulting in a 1.8 percentage point decrease in style consistency, a 1.8 percentage point drop in logical consistency, and an 11.3 percentage point reduction in entity consistency. Ablating entities causes a similarly catastrophic failure in entity consistency with a 10.6 percentage point decline. Excluding global style information leads to a significant 4.4 percentage point drop in style consistency while having smaller effects on logical and entity coherence. These findings indicate that the components of IUT-Plug are non-redundant, and each contributes uniquely to different aspects of multimodal image-text input and output tasks.

\subsection{Ablation on dynamic evaluation criteria}

The evaluation framework mentioned is dynamic and does not require predefined criteria, unlike methods relying on static metrics such as CLIP scores. We present the differences between our framework and existing static approaches in Table~\ref{tab:evaluator_comparison}. The static criterion baseline achieves only 55.3 percent agreement with human judgments when using our powerful evaluator model, Eval-3K. In contrast, our dynamic criterion approach generates context-specific questions, such as whether the cat is now sleeping on the red mat, and reaches 87.6 percent agreement with human annotators when used with Eval-3K. Static evaluation methods are either too broad to detect subtle attribute swaps or too narrow to handle novel compositional requests.

\subsection{Ablation on evaluation model's training data}

The amount and quality of data may affect model performance. We train and evaluate three variants to address this issue. One uses a small dataset of 500 expert-annotated samples (Eval-0.5K). Another uses 2,000 samples (Eval-2K). The full model is trained on 3,000 samples (Eval-3K). All models follow the same dynamic criterion generation process. Results are shown in Table~\ref{tab:evaluator_comparison}. They indicate a clear positive relationship between training data size and evaluation reliability. The model trained on 500 samples achieves 46.2\% agreement with human judgments. When scaled to 2,000 samples, agreement rises to 70.8\%. The full model reaches 87.6\% with 3,000 samples. This shows that human-labeled data is essential for strong evaluator performance. It also suggests room for further improvement with larger and higher-quality datasets.

\section{Conclusion}

We present \textbf{IUT-Plug}, a lightweight plug-in module for interleaved image-text generation. It mitigates multimodal context drift in logic, entity identity, and visual style. IUT-Plug operates between a frozen vision-language model and a text-to-image generator. It extracts a structured representation from the input image. This representation is called the Image Understanding Tree. The tree captures entities, their styles, and their relationships. The IUT-Plug is updated by textual instructions. It is then serialized into a json file for the downstream  text-to-image (T2I) model. This ensures that critical context and information are preserved across modalities. Our method requires no retraining of the base models.On the MMIE benchmark, IUT-Plug consistently improves consistency scores across all three dimensions. The gains range from 7.2 to 10.5 percentage points, with the largest improvement observed in entity consistency. These results confirm that explicit symbolic grounding can effectively bridge the consistency gap in modern multimodal pipelines.



\newpage

\bibliographystyle{plainnat}
\bibliography{iclr2026_conference_fix}

\begin{thebibliography}{31}
\providecommand{\natexlab}[1]{#1}
\providecommand{\url}[1]{\texttt{#1}}
\expandafter\ifx\csname urlstyle\endcsname\relax
  \providecommand{\doi}[1]{doi: #1}\else
  \providecommand{\doi}{doi: \begingroup \urlstyle{rm}\Url}\fi

\bibitem[Achiam et~al.(2023)Achiam, Adler, Agarwal, Ahmad, Akkaya, Aleman, Almeida, Altenschmidt, Altman, Anadkat, et~al.]{achiam2023gpt}
Josh Achiam, Steven Adler, Sandhini Agarwal, Lama Ahmad, Ilge Akkaya, Florencia~Leoni Aleman, Diogo Almeida, Janko Altenschmidt, Sam Altman, Shyamal Anadkat, et~al.
\newblock Gpt-4 technical report.
\newblock \emph{arXiv preprint arXiv:2303.08774}, 2023.

\bibitem[Alayrac et~al.(2022)Alayrac, Donahue, Luc, Miech, Barr, Hasson, Lenc, Mensch, Millican, Reynolds, et~al.]{alayrac2022flamingo}
Jean-Baptiste Alayrac, Jeff Donahue, Pauline Luc, Antoine Miech, Iain Barr, Yana Hasson, Karel Lenc, Arthur Mensch, Katherine Millican, Malcolm Reynolds, et~al.
\newblock Flamingo: a visual language model for few-shot learning.
\newblock \emph{Advances in neural information processing systems}, 35:\penalty0 23716--23736, 2022.

\bibitem[Anonymous(2025)]{anonymous2025why}
Anonymous.
\newblock Why settle for one? text-to-imageset generation and evaluation.
\newblock In \emph{Submitted to The Fourteenth International Conference on Learning Representations}, 2025.
\newblock URL \url{https://openreview.net/forum?id=0mBCl2goJr}.
\newblock under review.

\bibitem[Betker et~al.(2023)Betker, Goh, Jing, Brooks, Wang, Li, Ouyang, Zhuang, Lee, Guo, et~al.]{betker2023improving}
James Betker, Gabriel Goh, Li~Jing, Tim Brooks, Jianfeng Wang, Linjie Li, Long Ouyang, Juntang Zhuang, Joyce Lee, Yufei Guo, et~al.
\newblock Improving image generation with better captions.
\newblock \emph{Computer Science. https://cdn. openai. com/papers/dall-e-3. pdf}, 2\penalty0 (3):\penalty0 8, 2023.

\bibitem[Bougzime et~al.(2025)Bougzime, Jabbar, Cruz, and Demoly]{bougzime2025unlocking}
Oualid Bougzime, Samir Jabbar, Christophe Cruz, and Fr{\'e}d{\'e}ric Demoly.
\newblock Unlocking the potential of generative ai through neuro-symbolic architectures: Benefits and limitations.
\newblock \emph{arXiv preprint arXiv:2502.11269}, 2025.

\bibitem[Chang et~al.(2021)Chang, Ren, Xu, Li, Chen, and Hauptmann]{chang2021comprehensive}
Xiaojun Chang, Pengzhen Ren, Pengfei Xu, Zhihui Li, Xiaojiang Chen, and Alex Hauptmann.
\newblock A comprehensive survey of scene graphs: Generation and application.
\newblock \emph{IEEE Transactions on Pattern Analysis and Machine Intelligence}, 45\penalty0 (1):\penalty0 1--26, 2021.

\bibitem[Chen et~al.(2024)Chen, Zhao, Liu, Bai, Lin, Zhou, and Chang]{chen2024image}
Liang Chen, Haozhe Zhao, Tianyu Liu, Shuai Bai, Junyang Lin, Chang Zhou, and Baobao Chang.
\newblock An image is worth 1/2 tokens after layer 2: Plug-and-play inference acceleration for large vision-language models.
\newblock In \emph{European Conference on Computer Vision}, pages 19--35. Springer, 2024.

\bibitem[Christopher et~al.(2025)Christopher, Cardei, Liang, and Fioretto]{christopher2025neuro}
Jacob~K Christopher, Michael Cardei, Jinhao Liang, and Ferdinando Fioretto.
\newblock Neuro-symbolic generative diffusion models for physically grounded, robust, and safe generation.
\newblock \emph{arXiv preprint arXiv:2506.01121}, 2025.

\bibitem[d'Avila Garcez and Lamb(2020)]{d2020neurosymbolic}
Artur d'Avila Garcez and Luis~C Lamb.
\newblock Neurosymbolic ai: the 3rd wave.
\newblock \emph{arXiv e-prints}, pages arXiv--2012, 2020.

\bibitem[Esser et~al.(2024)Esser, Kulal, Blattmann, Entezari, M{\"u}ller, Saini, Levi, Lorenz, Sauer, Boesel, et~al.]{esser2024scaling}
Patrick Esser, Sumith Kulal, Andreas Blattmann, Rahim Entezari, Jonas M{\"u}ller, Harry Saini, Yam Levi, Dominik Lorenz, Axel Sauer, Frederic Boesel, et~al.
\newblock Scaling rectified flow transformers for high-resolution image synthesis.
\newblock In \emph{Forty-first international conference on machine learning}, 2024.

\bibitem[Johnson et~al.(2018)Johnson, Gupta, and Fei-Fei]{johnson2018image}
Justin Johnson, Agrim Gupta, and Li~Fei-Fei.
\newblock Image generation from scene graphs.
\newblock In \emph{Proceedings of the IEEE conference on computer vision and pattern recognition}, pages 1219--1228, 2018.

\bibitem[Krishna et~al.(2017)Krishna, Zhu, Groth, Johnson, Hata, Kravitz, Chen, Kalantidis, Li, Shamma, et~al.]{krishna2017visual}
Ranjay Krishna, Yuke Zhu, Oliver Groth, Justin Johnson, Kenji Hata, Joshua Kravitz, Stephanie Chen, Yannis Kalantidis, Li-Jia Li, David~A Shamma, et~al.
\newblock Visual genome: Connecting language and vision using crowdsourced dense image annotations.
\newblock \emph{International journal of computer vision}, 123\penalty0 (1):\penalty0 32--73, 2017.

\bibitem[Labs et~al.(2025)Labs, Batifol, Blattmann, Boesel, Consul, Diagne, Dockhorn, English, English, Esser, Kulal, Lacey, Levi, Li, Lorenz, Müller, Podell, Rombach, Saini, Sauer, and Smith]{labs2025flux1kontextflowmatching}
Black~Forest Labs, Stephen Batifol, Andreas Blattmann, Frederic Boesel, Saksham Consul, Cyril Diagne, Tim Dockhorn, Jack English, Zion English, Patrick Esser, Sumith Kulal, Kyle Lacey, Yam Levi, Cheng Li, Dominik Lorenz, Jonas Müller, Dustin Podell, Robin Rombach, Harry Saini, Axel Sauer, and Luke Smith.
\newblock Flux.1 kontext: Flow matching for in-context image generation and editing in latent space, 2025.
\newblock URL \url{https://arxiv.org/abs/2506.15742}.

\bibitem[Li et~al.(2024)Li, Li, Jing, Liu, Shao, Wu, Luo, Qiao, and Zhang]{li2024searchlvlms}
Chuanhao Li, Zhen Li, Chenchen Jing, Shuo Liu, Wenqi Shao, Yuwei Wu, Ping Luo, Yu~Qiao, and Kaipeng Zhang.
\newblock Searchlvlms: A plug-and-play framework for augmenting large vision-language models by searching up-to-date internet knowledge.
\newblock \emph{Advances in Neural Information Processing Systems}, 37:\penalty0 64582--64603, 2024.

\bibitem[Liu et~al.(2023)Liu, Li, Wu, and Lee]{liu2023visualinstructiontuning}
Haotian Liu, Chunyuan Li, Qingyang Wu, and Yong~Jae Lee.
\newblock Visual instruction tuning, 2023.
\newblock URL \url{https://arxiv.org/abs/2304.08485}.

\bibitem[Marcus(2020)]{marcus2020next}
Gary Marcus.
\newblock The next decade in ai: four steps towards robust artificial intelligence.
\newblock \emph{arXiv preprint arXiv:2002.06177}, 2020.

\bibitem[Meng et~al.(2024)Meng, Ma, Miao, Zhang, Yang, and Zhuang]{meng2024imageregenerationevaluatingtexttoimage}
Chutian Meng, Fan Ma, Jiaxu Miao, Chi Zhang, Yi~Yang, and Yueting Zhuang.
\newblock Image regeneration: Evaluating text-to-image model via generating identical image with multimodal large language models, 2024.
\newblock URL \url{https://arxiv.org/abs/2411.09449}.

\bibitem[Qwen et~al.(2025)Qwen, :, Yang, Yang, Zhang, Hui, Zheng, Yu, Li, Liu, Huang, Wei, Lin, Yang, Tu, Zhang, Yang, Yang, Zhou, Lin, Dang, Lu, Bao, Yang, Yu, Li, Xue, Zhang, Zhu, Men, Lin, Li, Tang, Xia, Ren, Ren, Fan, Su, Zhang, Wan, Liu, Cui, Zhang, and Qiu]{qwen2025qwen25technicalreport}
Qwen, :, An~Yang, Baosong Yang, Beichen Zhang, Binyuan Hui, Bo~Zheng, Bowen Yu, Chengyuan Li, Dayiheng Liu, Fei Huang, Haoran Wei, Huan Lin, Jian Yang, Jianhong Tu, Jianwei Zhang, Jianxin Yang, Jiaxi Yang, Jingren Zhou, Junyang Lin, Kai Dang, Keming Lu, Keqin Bao, Kexin Yang, Le~Yu, Mei Li, Mingfeng Xue, Pei Zhang, Qin Zhu, Rui Men, Runji Lin, Tianhao Li, Tianyi Tang, Tingyu Xia, Xingzhang Ren, Xuancheng Ren, Yang Fan, Yang Su, Yichang Zhang, Yu~Wan, Yuqiong Liu, Zeyu Cui, Zhenru Zhang, and Zihan Qiu.
\newblock Qwen2.5 technical report, 2025.
\newblock URL \url{https://arxiv.org/abs/2412.15115}.

\bibitem[Ramesh et~al.(2022)Ramesh, Dhariwal, Nichol, Chu, and Chen]{ramesh2022hierarchical}
Aditya Ramesh, Prafulla Dhariwal, Alex Nichol, Casey Chu, and Mark Chen.
\newblock Hierarchical text-conditional image generation with clip latents.
\newblock \emph{arXiv preprint arXiv:2204.06125}, 2022.

\bibitem[Rombach et~al.(2022)Rombach, Blattmann, Lorenz, Esser, and Ommer]{rombach2022high}
Robin Rombach, Andreas Blattmann, Dominik Lorenz, Patrick Esser, and Bj{\"o}rn Ommer.
\newblock High-resolution image synthesis with latent diffusion models.
\newblock In \emph{Proceedings of the IEEE/CVF conference on computer vision and pattern recognition}, pages 10684--10695, 2022.

\bibitem[Sarker et~al.(2021)Sarker, Zhou, Eberhart, and Hitzler]{sarker2021neuro}
Md~Kamruzzaman Sarker, Lu~Zhou, Aaron Eberhart, and Pascal Hitzler.
\newblock Neuro-symbolic artificial intelligence: Current trends, 2021.
\newblock URL \url{https://arxiv.org/abs/2105.05330}.

\bibitem[Sun et~al.(2024)Sun, Cui, Zhang, Zhang, Yu, Wang, Rao, Liu, Huang, and Wang]{sun2024generative}
Quan Sun, Yufeng Cui, Xiaosong Zhang, Fan Zhang, Qiying Yu, Yueze Wang, Yongming Rao, Jingjing Liu, Tiejun Huang, and Xinlong Wang.
\newblock Generative multimodal models are in-context learners.
\newblock In \emph{Proceedings of the IEEE/CVF Conference on Computer Vision and Pattern Recognition}, pages 14398--14409, 2024.

\bibitem[Team et~al.(2024)Team, Georgiev, Lei, Burnell, Bai, Gulati, Tanzer, Vincent, Pan, Wang, et~al.]{team2024gemini}
Gemini Team, Petko Georgiev, Ving~Ian Lei, Ryan Burnell, Libin Bai, Anmol Gulati, Garrett Tanzer, Damien Vincent, Zhufeng Pan, Shibo Wang, et~al.
\newblock Gemini 1.5: Unlocking multimodal understanding across millions of tokens of context.
\newblock \emph{arXiv preprint arXiv:2403.05530}, 2024.

\bibitem[Tian et~al.(2024)Tian, Zhu, Xiong, Wang, Chen, Wang, Chen, Lu, Lu, Zhou, et~al.]{tian2024mm}
Changyao Tian, Xizhou Zhu, Yuwen Xiong, Weiyun Wang, Zhe Chen, Wenhai Wang, Yuntao Chen, Lewei Lu, Tong Lu, Jie Zhou, et~al.
\newblock Mm-interleaved: Interleaved image-text generative modeling via multi-modal feature synchronizer.
\newblock \emph{arXiv preprint arXiv:2401.10208}, 2024.

\bibitem[Wang et~al.(2024)Wang, Bai, Tan, Wang, Fan, Bai, Chen, Liu, Wang, Ge, et~al.]{wang2024qwen2}
Peng Wang, Shuai Bai, Sinan Tan, Shijie Wang, Zhihao Fan, Jinze Bai, Keqin Chen, Xuejing Liu, Jialin Wang, Wenbin Ge, et~al.
\newblock Qwen2-vl: Enhancing vision-language model's perception of the world at any resolution.
\newblock \emph{arXiv preprint arXiv:2409.12191}, 2024.

\bibitem[Wu et~al.(2024)Wu, Fei, Qu, Ji, and Chua]{wu2024next}
Shengqiong Wu, Hao Fei, Leigang Qu, Wei Ji, and Tat-Seng Chua.
\newblock Next-gpt: Any-to-any multimodal llm.
\newblock In \emph{Forty-first International Conference on Machine Learning}, 2024.

\bibitem[Xia et~al.(2024)Xia, Han, Qiu, Zhou, Wang, Zheng, Chen, Cui, Ding, Li, et~al.]{xia2024mmie}
Peng Xia, Siwei Han, Shi Qiu, Yiyang Zhou, Zhaoyang Wang, Wenhao Zheng, Zhaorun Chen, Chenhang Cui, Mingyu Ding, Linjie Li, et~al.
\newblock Mmie: Massive multimodal interleaved comprehension benchmark for large vision-language models.
\newblock \emph{arXiv preprint arXiv:2410.10139}, 2024.

\bibitem[Yang et~al.(2022)Yang, Huang, Song, Hong, Li, Zhang, Cui, Ghanem, and Yang]{yang2022diffusion}
Ling Yang, Zhilin Huang, Yang Song, Shenda Hong, Guohao Li, Wentao Zhang, Bin Cui, Bernard Ghanem, and Ming-Hsuan Yang.
\newblock Diffusion-based scene graph to image generation with masked contrastive pre-training.
\newblock \emph{arXiv preprint arXiv:2211.11138}, 2022.

\bibitem[Zhang et~al.(2023)Zhang, Zhang, Vineet, Joshi, and Wang]{zhang2023controllable}
Tianjun Zhang, Yi~Zhang, Vibhav Vineet, Neel Joshi, and Xin Wang.
\newblock Controllable text-to-image generation with gpt-4.
\newblock \emph{arXiv preprint arXiv:2305.18583}, 2023.

\bibitem[Zheng et~al.(2023)Zheng, He, and Wang]{zheng2023minigpt}
Kaizhi Zheng, Xuehai He, and Xin~Eric Wang.
\newblock Minigpt-5: Interleaved vision-and-language generation via generative vokens.
\newblock \emph{arXiv preprint arXiv:2310.02239}, 2023.

\bibitem[Zhou et~al.(2025)Zhou, Peng, Song, Li, Xu, Yang, Guo, Zhang, Lin, He, Zhao, Liu, Li, Xie, Chang, Qiao, Shao, and Zhang]{Zhou_2025_OPENING}
Pengfei Zhou, Xiaopeng Peng, Jiajun Song, Chuanhao Li, Zhaopan Xu, Yue Yang, Ziyao Guo, Hao Zhang, Yuqi Lin, Yefei He, Lirui Zhao, Shuo Liu, Tianhua Li, Yuxuan Xie, Xiaojun Chang, Yu~Qiao, Wenqi Shao, and Kaipeng Zhang.
\newblock Opening: A comprehensive benchmark for judging open-ended interleaved image-text generation.
\newblock In \emph{Proceedings of the {IEEE/CVF} Conference on Computer Vision and Pattern Recognition ({CVPR})}, pages 56--66, 2025.

\end{thebibliography}

\appendix

\newpage

\section{Appendix}

\subsection{Structured Text Prompting (STP) Baseline}
\label{sec:stp_baseline}

The Structured Text Prompting (STP) baseline is introduced to rigorously validate the necessity of IUT-Plug's \textbf{persistent, incrementally updated symbolic state} over merely using structured instructions within a single VLM turn. STP simulates an enhanced, state-of-the-art VLM capable of following complex formatting rules, yet lacks an external, memory-augmented reasoning module.

In the STP method, the base Vision-Language Model (VLM) is given the full history $(I_0, Q_1, \dots, Q_t)$ and is explicitly instructed to generate its textual response $T_t$ followed by a structured, self-contained summary of the current scene state $S_t$. However, this state $S_t$ is not programmatically parsed, maintained, or updated by an external module; instead, the VLM must regenerate the entire description from scratch in the next turn, relying solely on its internal attention mechanisms to maintain long-term context.

\textbf{Prompt template for the VLM in the STP baseline}. (using $T_{t-1}$ as the previous model response and $Q_t$ as the current user instruction)

\begin{quote}
\centering
\fbox{\parbox{0.95\linewidth}{
\textbf{Instruction:} Based on the image and the dialogue history provided, first generate your natural language response $T_t$. Following your response, you must generate a full JSON description of the current visual scene state, $S_t$. This JSON must list all entities, their attributes, and their relationships.

\textbf{JSON Schema (Must Follow):}\\
\scriptsize
\texttt{\{}\\
\texttt{~~"scene\_summary": "a short textual summary of the scene.",}\\
\texttt{~~"style": \{}\\
\texttt{~~~~"artistic\_medium": "...",}\\
\texttt{~~~~"lighting": "...",}\\
\texttt{~~~~"...": "..."}\\
\texttt{~~\},}\\
\texttt{~~"objects": [}\\
\texttt{~~~~\{"name": "...", "attributes": "...", "relationships": "... "\}}\\
\texttt{~~~~// ... all other key entities}\\
\texttt{~~],}\\
\texttt{~~"relationships": [}\\
\texttt{~~~~"entity\_A [relation] entity\_B",}\\
\texttt{~~~~// ... all key relations}\\
\texttt{~~]}\\
\texttt{\}}
\normalsize

\textbf{History and Input:} \\
\texttt{[Previous Turns and Image History]} \\
\texttt{Current Instruction ($Q_t$): [User Input $Q_t$]}
}}
\end{quote}

The T2I model for STP is then prompted using the VLM's natural language response $T_t$ concatenated with the VLM-generated JSON summary $S_t$, similar to how the full IUT-Plug module operates. The experimental results in Section 4.2 confirm that while STP offers an improvement over the vanilla baseline (proving that structured information helps), it consistently falls short of the IUT-Plug, demonstrating the critical role of the IUT's \textbf{explicit, incremental state update mechanism} in mitigating multimodal context drift.

\subsection{Key Components and Evaluation Criteria}

The core of our methodology relies on two key components. The first is the \texttt{construct\_IUT()} function, which parses visual scenes into a hierarchical structure, including object relationships, attributes for each node, and cross-object spatial relations. The second is the \texttt{GPT4v\_score()} function, which computes the structural consistency of generated images using a weighted combination of CLIP, DINO, and IUT alignment scores, as defined by the formula:
\[
\gamma = \alpha \cdot \text{CLIP}(I,I_{ref}) + \beta \cdot \text{DINO}(I,I_{ref}) + \lambda \cdot \text{IUT\_Alignment}(I,T)
\]
where $\lambda$ weights the IUT-based consistency verification.

For evaluation, we assess the generated content against the following key criteria:
\begin{enumerate}
    \item \textbf{Correctness}: Factual accuracy and validity of the content.
    \item \textbf{Image-Text Coherency}: The degree of alignment between visual and textual elements.
    \item \textbf{Multi-Step Consistency}: Thematic and stylistic consistency across multiple generation steps.
    \item \textbf{Content Quality}: The clarity of images and fluency of the text.
    \item \textbf{Completeness}: Ensuring no required information or steps are omitted in the output.
    \item \textbf{Content Richness}: The diversity and depth of the generated content.
\end{enumerate}

\subsection{Overview of Baseline Models}
The following models were used as baselines in our experiments:
\begin{itemize}
    \item \textbf{MiniGPT} \citep{zheng2023minigpt}: Combines MiniGPT-4 and Stable Diffusion for multimodal I/O, using "generative tokens" to bridge text and vision.
    \item \textbf{EMU-2} \citep{sun2024generative}: A 37B parameter generative multimodal model. We use a pipeline of its chat (Emu2-Chat) and generation (Emu2-Gen) variants.
    \item \textbf{GPT-4o} \citep{achiam2023gpt}: An advanced multimodal model from OpenAI capable of processing both text and visual inputs.
    \item \textbf{Gemini-1.5} \citep{team2024gemini}: A large language model from Google AI trained on a massive text and code dataset, with capabilities for image analysis.
    \item \textbf{LLaVA-34b} \citep{liu2023visualinstructiontuning}: An end-to-end model connecting a vision encoder with the Hermes-Yi-34B LLM for visual and language understanding. It does not support multiple image inputs.
    \item \textbf{Qwen2-VL-72b} \citep{wang2024qwen2}: The multimodal version of Alibaba Cloud's Qwen large model series, designed for text, image, and audio processing.
    \item \textbf{Openjourney}: A Stable Diffusion variant fine-tuned on Midjourney images for artistic and creative image generation.
    \item \textbf{Stable Diffusion 3 Medium} \citep{esser2024scaling}: A text-to-image model from Stability AI that generates high-quality images with fine detail.
    \item \textbf{Stable Diffusion XL turbo} \citep{esser2024scaling}: An optimized version of SDXL for accelerated, high-quality image generation.
    \item \textbf{Flux.1-dev}: A 12B parameter rectified flow transformer model from Black Forest Labs for efficient text-to-image and image-to-image tasks.
\end{itemize}

\subsection{Prompts for Image Generation Models}
This section details the two main prompt templates used to instruct the Large Language Model (LLM) for generating image prompts, both with and without the guidance of the Image Understanding Tree (IUT).

\paragraph{Prompt for LLM with IUT guidance.}
\label{lst:prompt_with_iut}
\begin{quote}
\small
\noindent
\texttt{%
Based on the text provided by the user (containing \#\#\#Question: and \#\#\#Answer: sections, where \#\#\#Question includes question text and \#\#\#Description of image), generate detailed descriptive prompts for an image generation model to explain the \#\#\#Answer section: \\
1. Note that you should decide how many image prompts to generate, but no more than 2 image prompts; \\
2. Each image prompt should be clearly differentiated from others if the description refers to the same scene, include it in a single image prompt (determining whether it's the same scene should be analyzed in conjunction with the \#\#\#Description of image); descriptions for different images should represent distinct scenes. However, if there is a clear sequence or step-by-step process in the \#\#\#Answer, different images can represent different steps, but each image description should still be as detailed as possible; \\
3. Each image description should be detailed and descriptive, suitable for image generation; \\
4. Note that each image prompt must begin with <image>, do not add any extra text, explanations, or numbering. Output only the prompts separated by <image> tags. For example: <image>A whimsical outdoor Halloween patio scene... <image>A third individual seated...; \\
5. When generating prompts, give priority to referring to the \#\#\#Description of image section in the \#\#\#Question; this helps maintain consistency in style and content between the images in the question and answer; \\
6. Regarding the \#\#\#Answer section: \\
a. If the current \#\#\#Answer provided by the user contains <image> and </image> tags, prioritize understanding the text between these tags and combine it with the \#\#\#Description of image to create new detailed prompts; \\
b. If multiple pairs of <image> and </image> tags exist in the \#\#\#Answer, assess the relevance of the content if related, merge them into one image description if unrelated, separate them; if the original \#\#\#Answer describes a step-by-step process, different steps can be represented in different image descriptions; \\
c. If no <image> and </image> tags are present in the \#\#\#Answer, or if alternative markers such as (image) or (/image) are used, analyze the surrounding text and combine it with the \#\#\#Description of image to generate detailed prompts.%
}
\end{quote}

\paragraph{Prompt for LLM without IUT guidance.}
\label{lst:prompt_without_iut}
\begin{quote}
\small
\noindent
\texttt{%
Based on the text provided by the user, generate a series of descriptive prompts for an image generation model: \\
1. Note, generate only 1 or 2 sets of prompts, no more than 2 sets; \\
2. Note, each set of prompts should be between 50 and 200 characters in length; \\
3. If the current user-provided text contains <image> and </image> tags, prioritize recognizing the text between these tags; \\
4. If there are multiple pairs of <image> and </image> tags, generate multiple sets of prompts; \\
5. If there are no pairs of <image> and </image> tags or if there are similar tags such as (image) or (/image) besides the tag pairs, also analyze the nearby text to generate prompts; \\
6. Each prompt should correspond to different visual elements or scenes described in the text; \\
7. Note, begin each image's prompt with <image>, without adding any additional text, explanations, or numbering. Only output content separated by <image>. For example: <image>A majestic mountain range at sunrise. <image>A serene lake reflecting the colorful sky; A dense forest with tall pine trees.%
}
\end{quote}

\subsection{Illustrative Cases}
As shown in the figures below, we select several examples from various categories for demonstration, including the input questions (both images and text) and the outputs of the evaluated models.

\subsubsection{IUT Performance Examples}
This section provides qualitative examples comparing the outputs of the interleaved generation task with and without the IUT-Plug.

\begin{figure}[t!]
    \centering
    \begin{subfigure}{0.32\textwidth}
        \includegraphics[width=\linewidth]{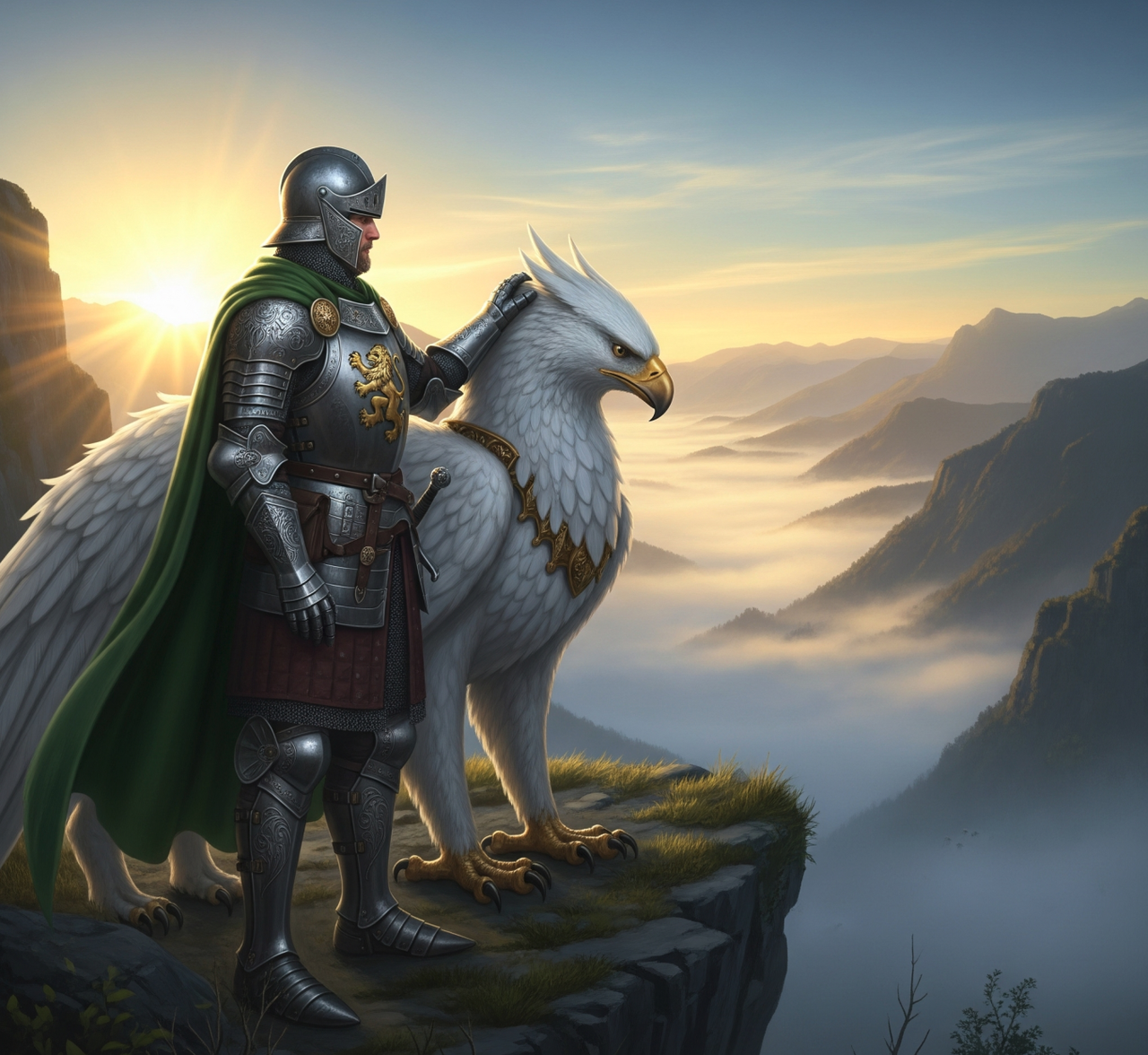}
        \caption{Initial Image for Q1.}
    \end{subfigure}
    \hfill
    \begin{subfigure}{0.32\textwidth}
        \includegraphics[width=\linewidth]{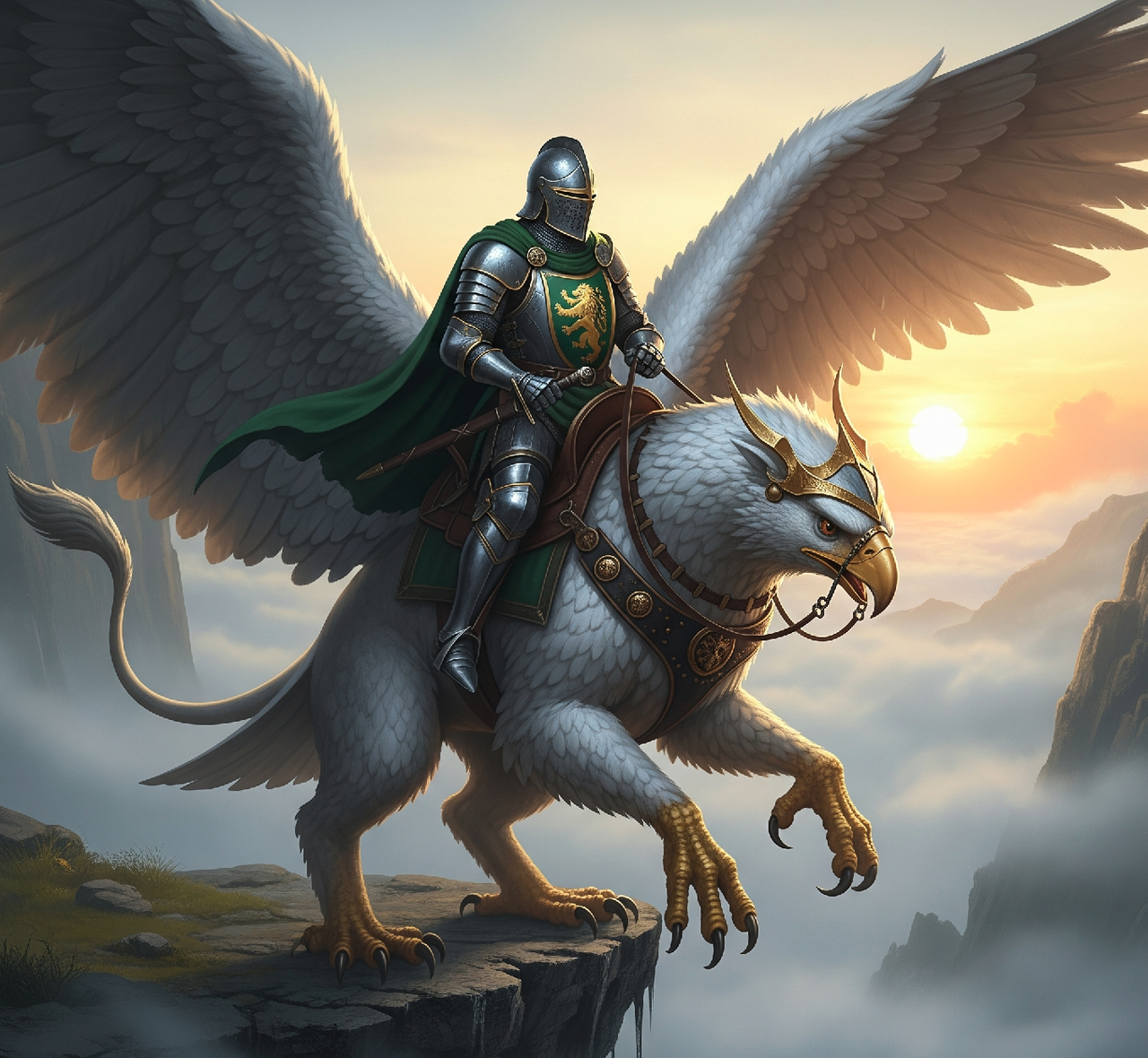}
        \caption{Generated with IUT.}
    \end{subfigure}
    \hfill
    \begin{subfigure}{0.32\textwidth}
        \includegraphics[width=\linewidth]{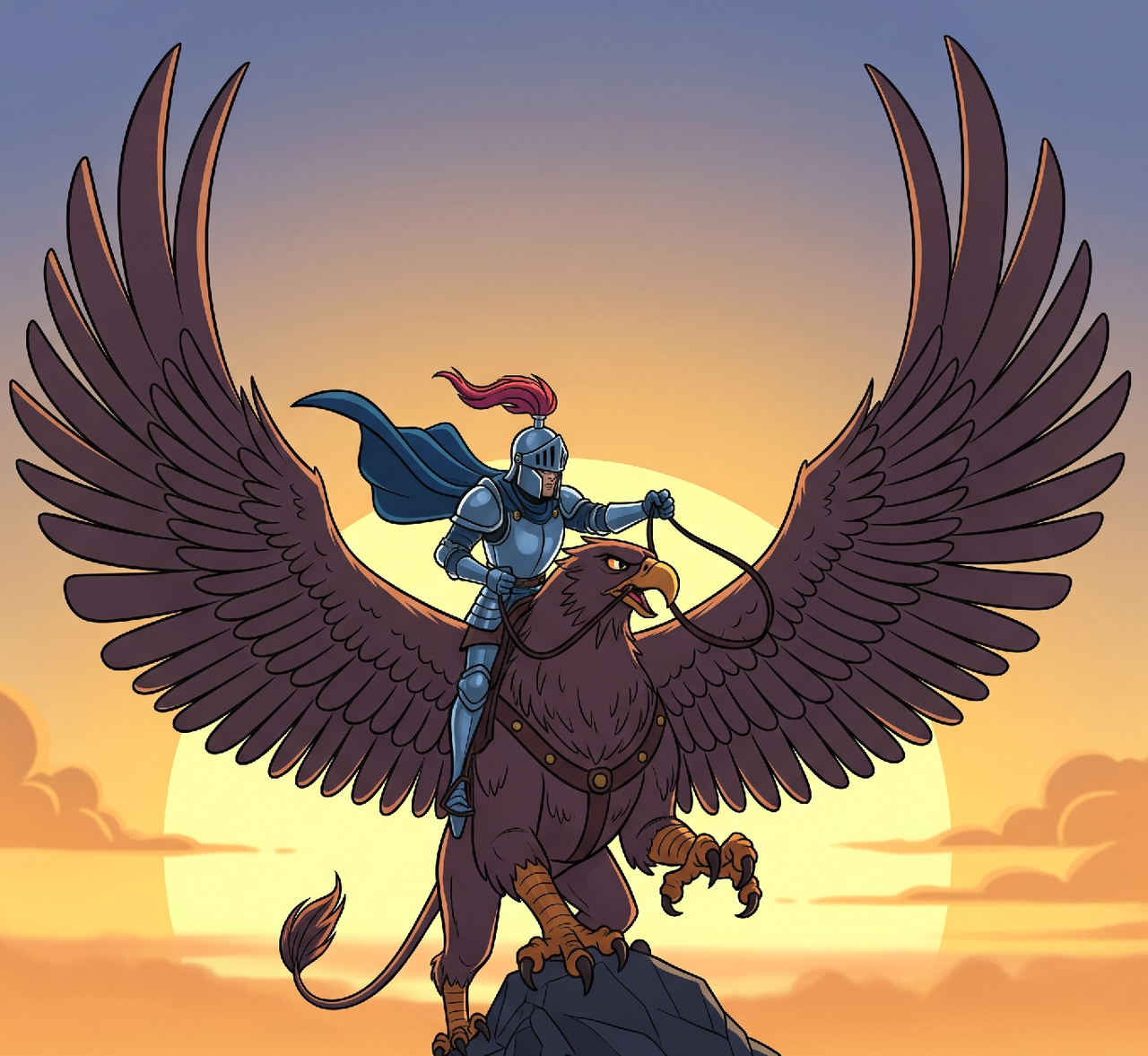}
        \caption{Generated without IUT.}
    \end{subfigure}
    \caption{Example 1. \textbf{Q:} A knight and his griffin companion prepare to set off at dawn. \textbf{A:} The knight mounted his griffin, which spread its massive wings, ready to take flight towards the rising sun, its posture full of power.}
    \label{fig:case1_knight_appendix}
\end{figure}

\begin{figure}[t!]
    \centering
    \begin{subfigure}{0.32\textwidth}
        \includegraphics[width=\linewidth]{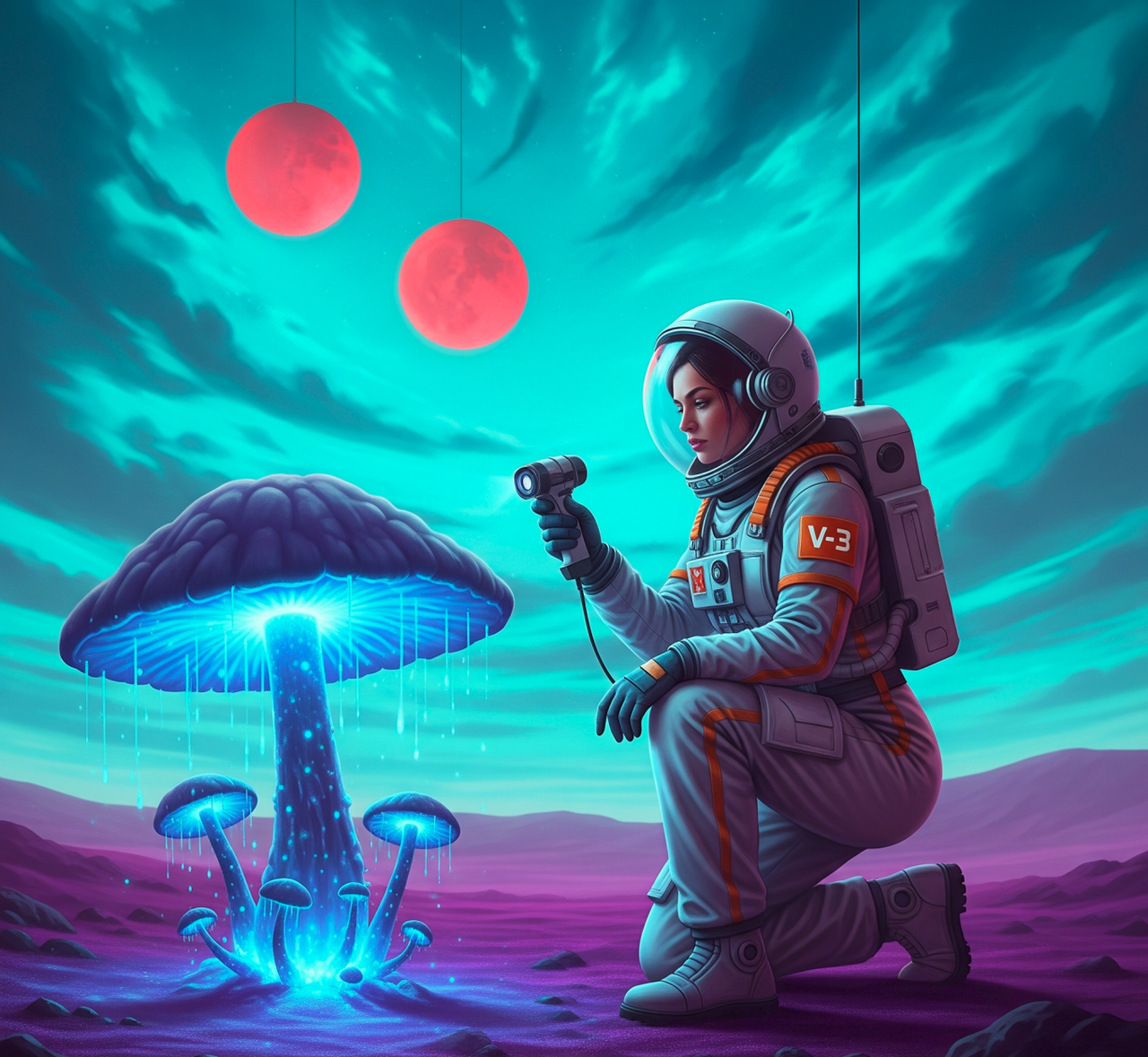}
        \caption{Initial Image for Q2.}
    \end{subfigure}
    \hfill
    \begin{subfigure}{0.32\textwidth}
        \includegraphics[width=\linewidth]{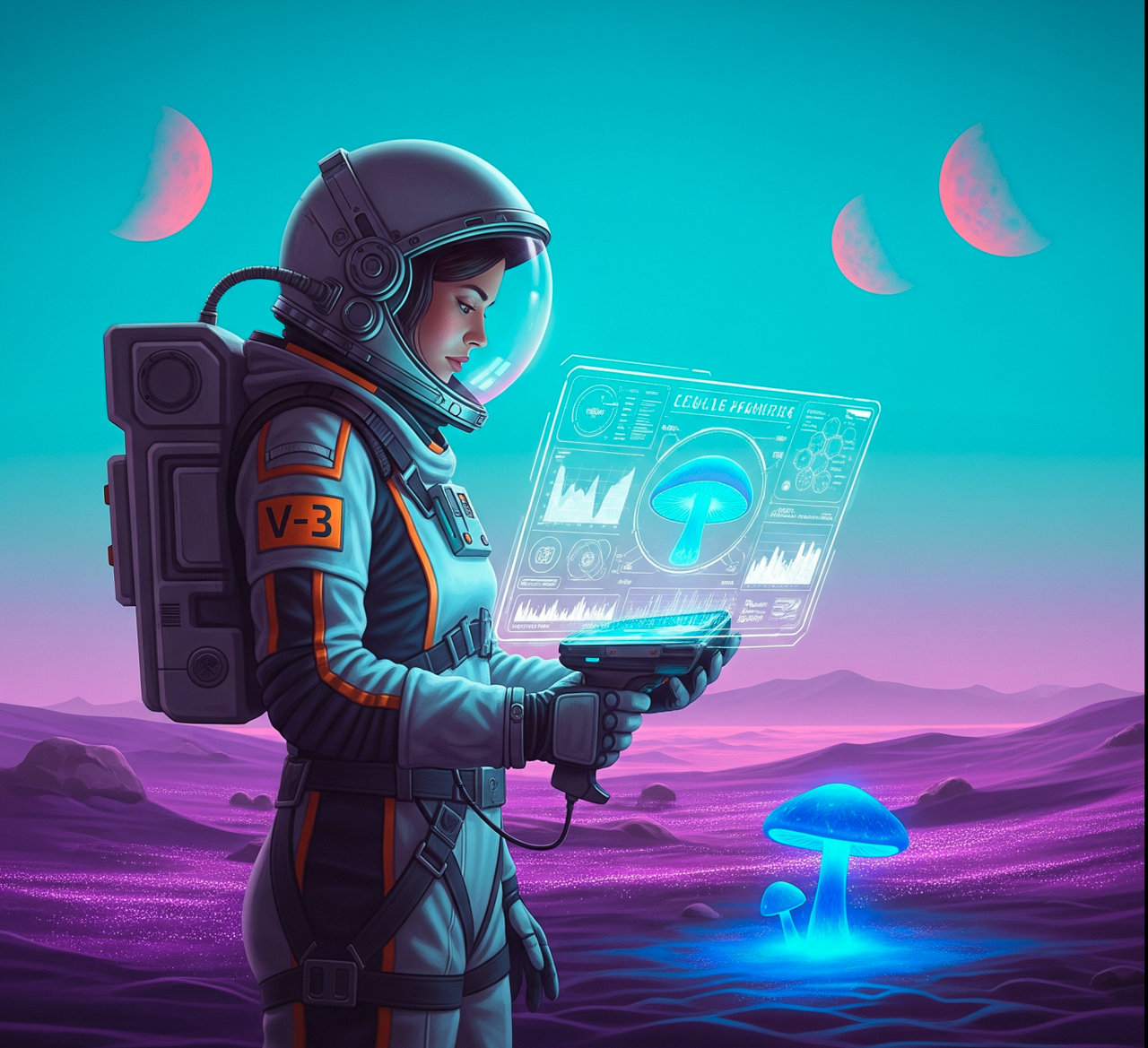}
        \caption{Generated with IUT.}
    \end{subfigure}
    \hfill
    \begin{subfigure}{0.32\textwidth}
        \includegraphics[width=\linewidth]{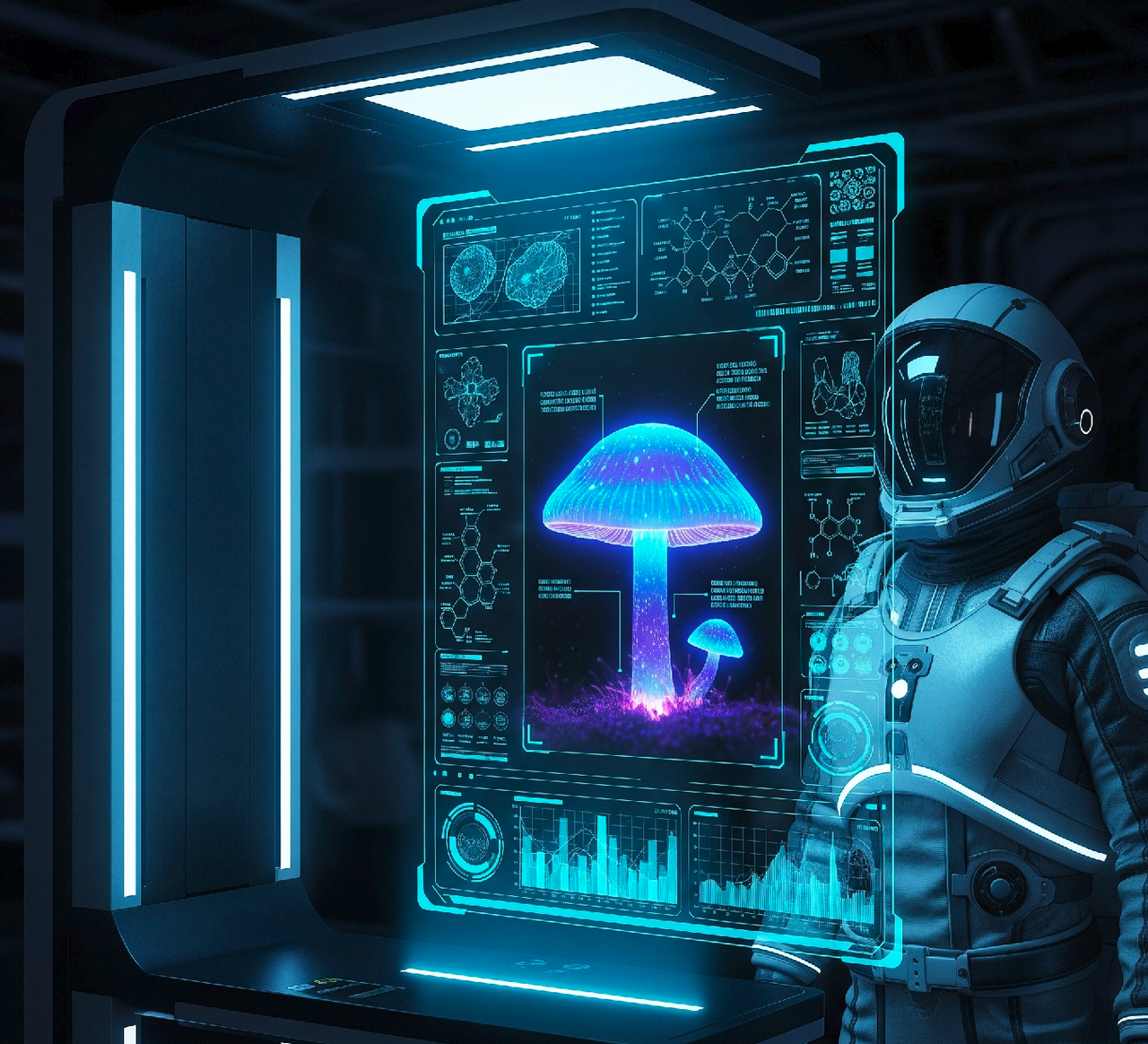}
        \caption{Generated without IUT.}
    \end{subfigure}
    \caption{Example 2. \textbf{Q:} An astronaut discovers glowing plants on an alien planet. \textbf{A:} The astronaut stood up, and the scanner in front of her projected a translucent holographic screen displaying complex data about the glowing mushroom.}
    \label{fig:case2_astronaut_appendix}
\end{figure}

\begin{figure}[t!]
    \centering
    \begin{subfigure}{0.32\textwidth}
        \includegraphics[width=\linewidth]{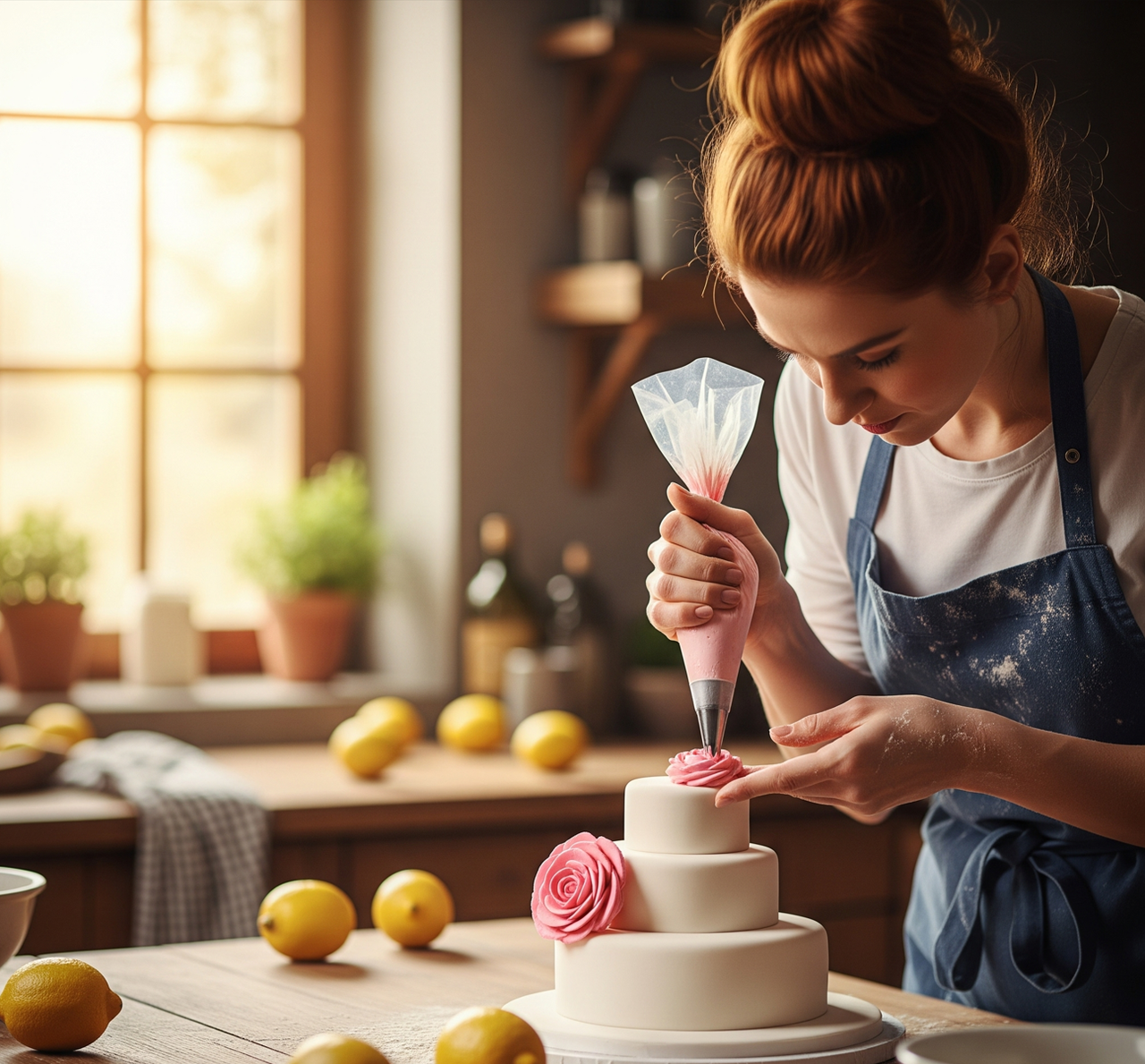}
        \caption{Initial Image for Q3.}
    \end{subfigure}
    \hfill
    \begin{subfigure}{0.32\textwidth}
        \includegraphics[width=\linewidth]{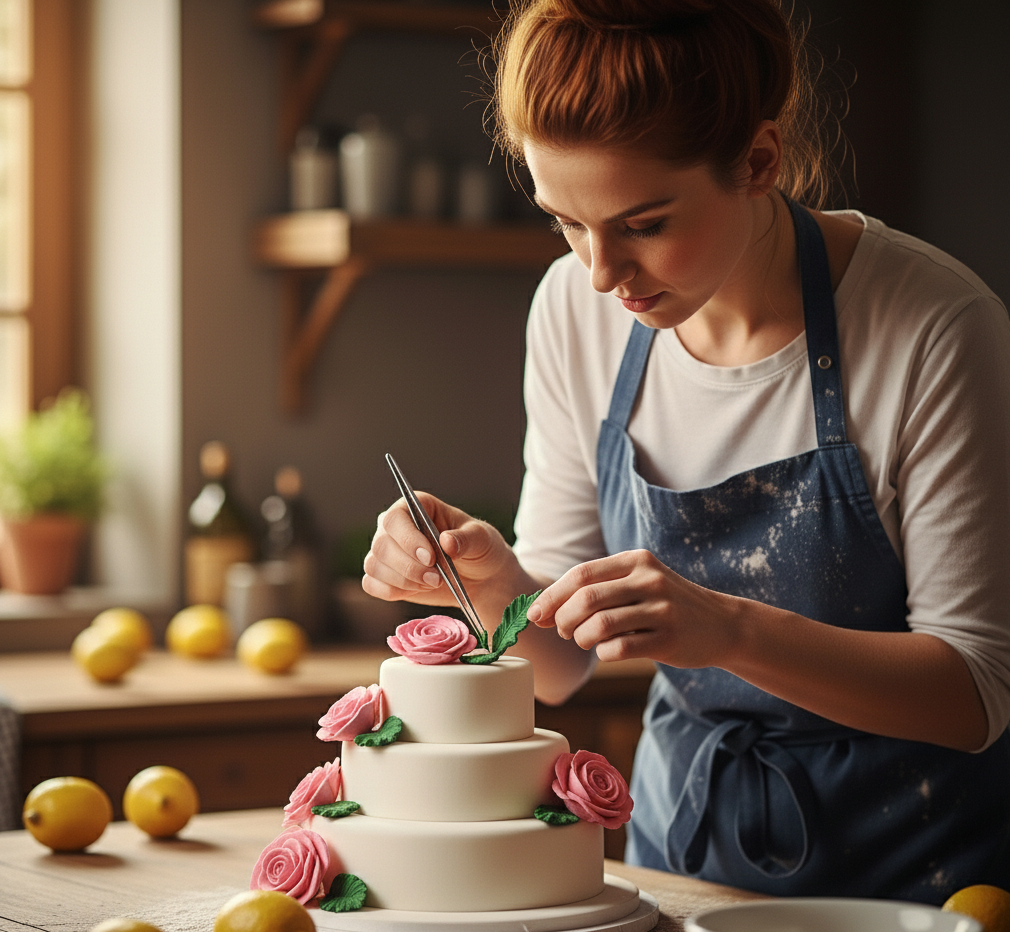}
        \caption{Generated with IUT.}
    \end{subfigure}
    \hfill
    \begin{subfigure}{0.32\textwidth}
        \includegraphics[width=\linewidth]{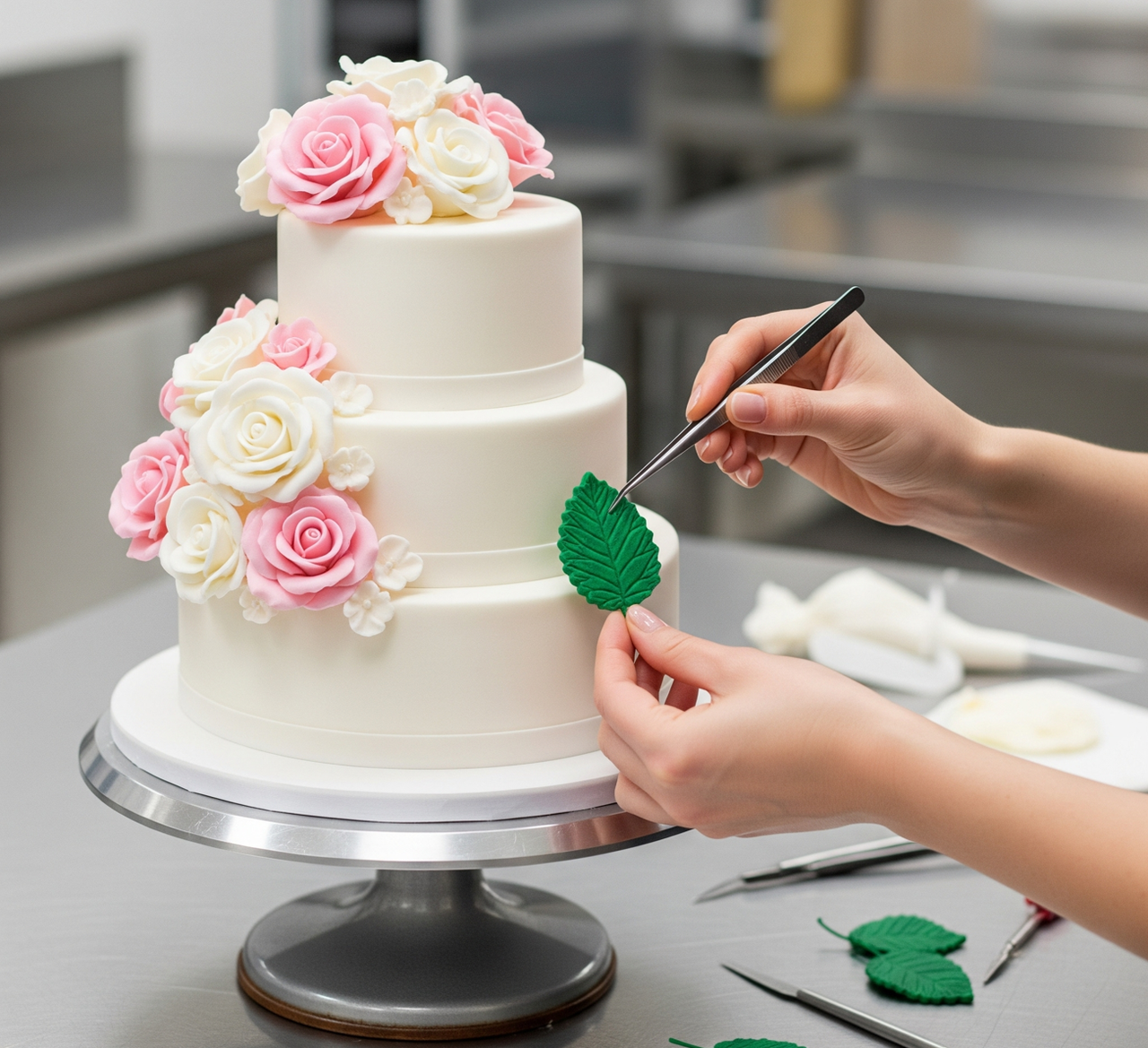}
        \caption{Generated without IUT.}
    \end{subfigure}
    \caption{Example 3. \textbf{Q:} A female baker is focused on decorating a three-tiered cake. \textbf{A:} The three-tiered white cake is now adorned with several roses. She is using tweezers to place a frosting leaf next to a rose, nearing the completion of the cake's decoration.}
    \label{fig:case3_baker_appendix}
\end{figure}

\subsubsection{Ablation Study Examples}
These images support the ablation study, showing the results when key components of the IUT structure (entity, relation, style) are omitted during generation.

\begin{figure}[t!]
    \centering
    \begin{subfigure}{0.32\textwidth}
        \includegraphics[width=\linewidth]{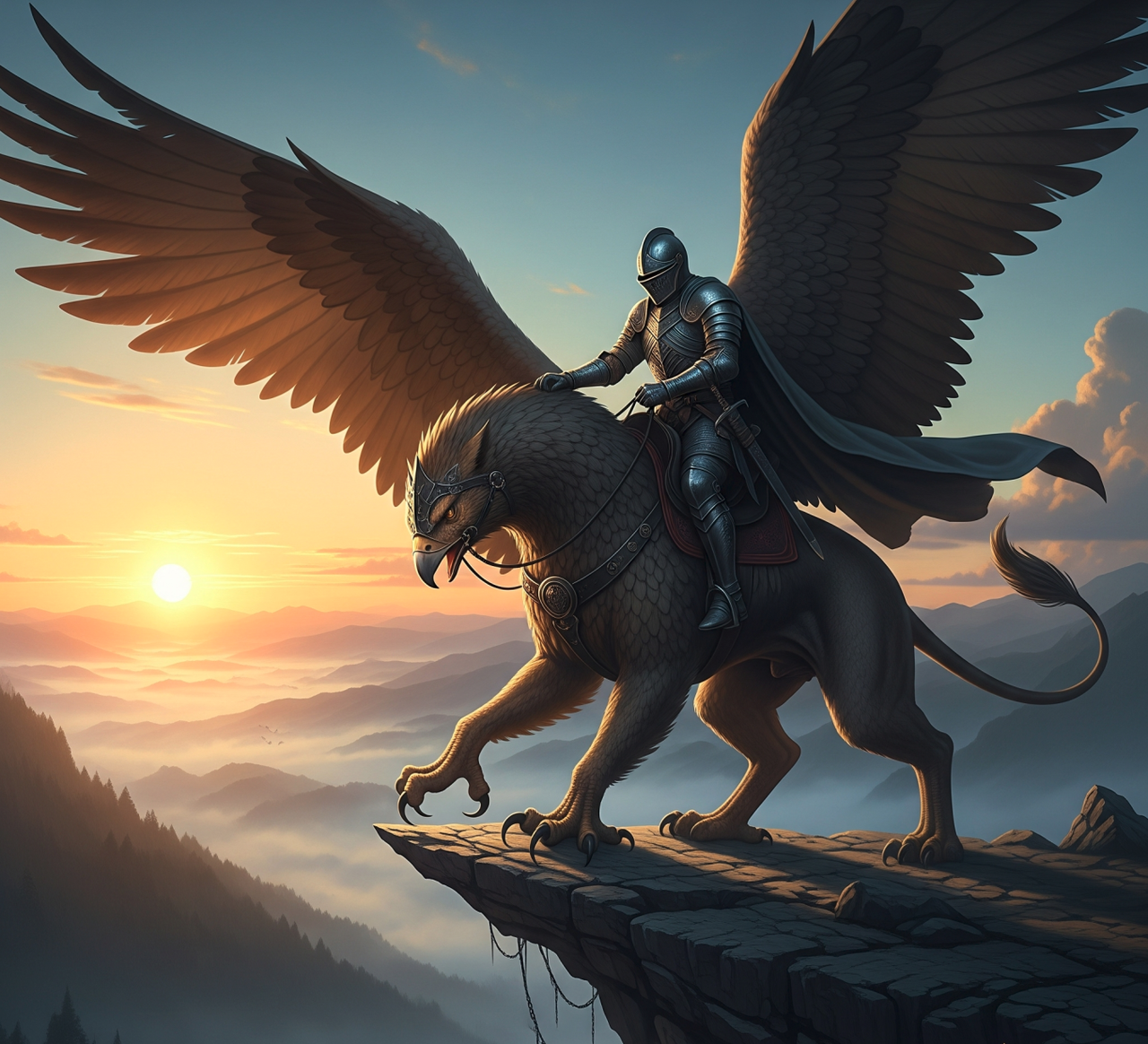}
        \caption{Generated w/o Entity.}
    \end{subfigure}
    \hfill
    \begin{subfigure}{0.32\textwidth}
        \includegraphics[width=\linewidth]{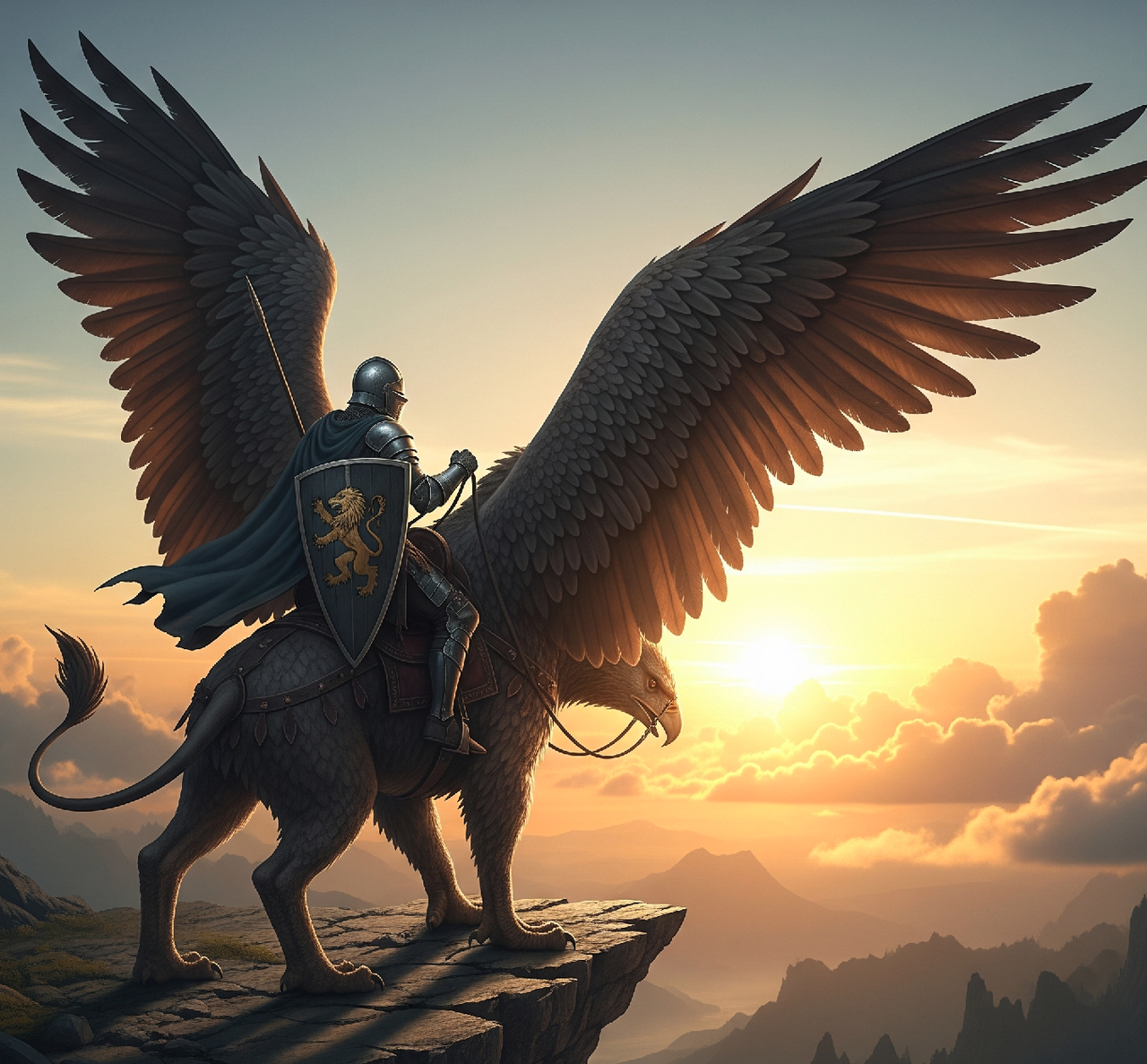}
        \caption{Generated w/o Relation.}
    \end{subfigure}
    \hfill
    \begin{subfigure}{0.32\textwidth}
        \includegraphics[width=\linewidth]{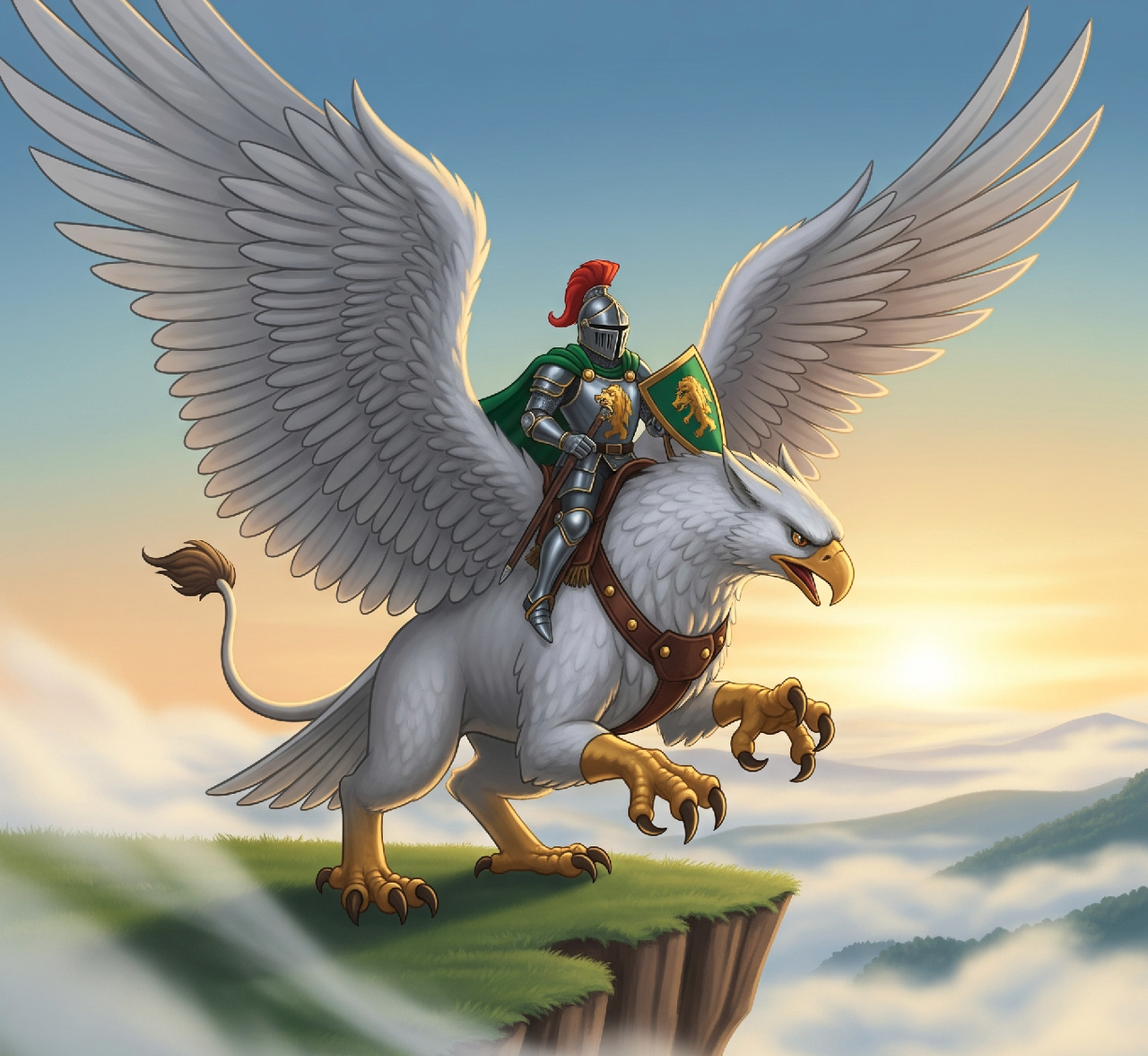}
        \caption{Generated w/o Style.}
    \end{subfigure}
    \caption{Ablation study for the knight and griffin example. The images show outputs when entity, relation, or style information is omitted from the IUT guidance.}
    \label{fig:ablation_knight_appendix}
\end{figure}

\subsubsection{IUT Extraction Examples}
This section provides concrete examples of the structured JSON output generated by the IUT extraction module for given images.

\begin{figure}[t!]
    \centering
    \includegraphics[width=0.7\linewidth]{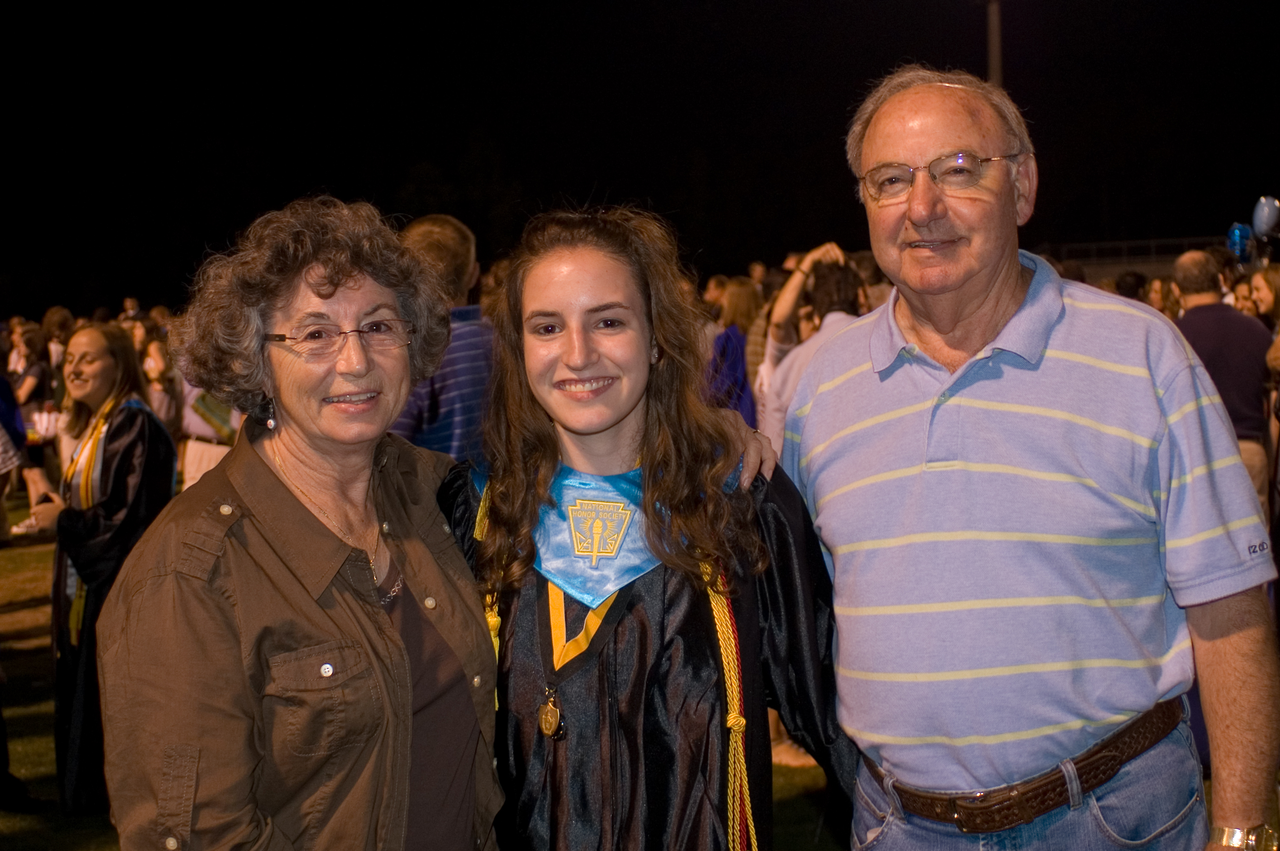}
    \caption{Input image for the first IUT extraction example (Graduation photo).}
    \label{fig:iut_input1_appendix}
\end{figure}

\paragraph{IUT JSON output for the graduation photo.}
\label{lst:iut_example1_appendix}
\begin{quote}
\small
\noindent
\texttt{%
\{ \\
\ \ "global\_description": "The primary subject of a family posing for a photo at a graduation event features a young woman in a blue and gold sash, flanked by an older couple in casual attire under artificial lighting, captured in a realistic style with warm tones, evoking a sense of pride and nostalgia.", \\
\ \ "global\_features": \{ \\
\ \ \ \ "style": "photorealistic", \\
\ \ \ \ "lighting": "soft artificial light", \\
\ \ \ \ "...": "..." \\
\ \ \}, \\
\ \ "objects": [ \\
\ \ \ \{"name": "woman wearing glasses", "type": "person", "...": "..." \}, \\
\ \ \ \{"name": "graduate in cap and gown", "type": "person", "...": "..." \} \\
\ \ ], \\
\ \ "relationships": [ \\
\ \ \ \ "woman standing next to graduate", \\
\ \ \ \ "man standing next to graduate", \\
\ \ \ \ "..." \\
\ \ ] \\
\}%
}
\end{quote}
\begin{figure}[t!]
    \centering
    \includegraphics[width=0.7\linewidth]{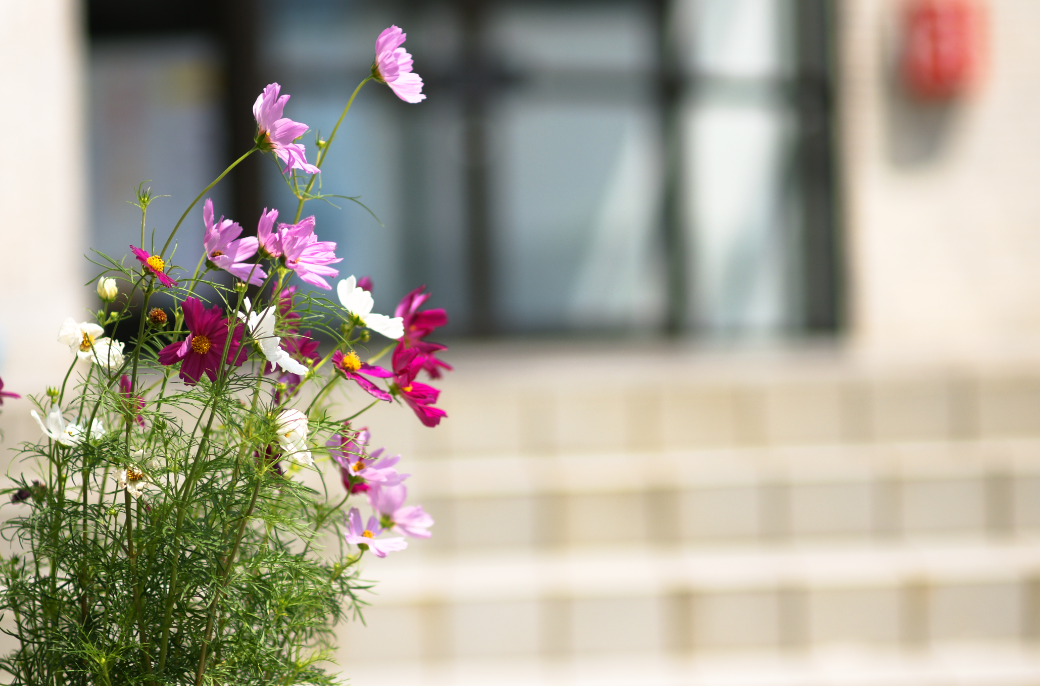}
    \caption{Input image for the second IUT extraction example (Flowers).}
    \label{fig:iut_input2_appendix}
\end{figure}

\paragraph{IUT JSON output for the flowers photo.}
\label{lst:iut_example2_appendix}
\begin{quote}
\small
\noindent
\texttt{%
\{ \\
\ \ "global\_description": "The vibrant bouquet of cosmos flowers in shades of pink, purple, and white, set against a softly blurred urban backdrop, showcases a realistic and detailed artistic style that evokes a serene and peaceful atmosphere.", \\
\ \ "global\_features": \{ \\
\ \ \ \ "style": "photorealistic", \\
\ \ \ \ "lighting": "soft natural light", \\
\ \ \ \ "...": "..." \\
\ \ \}, \\
\ \ "objects": [ \\
\ \ \ \{"name": "colorful flowers", "type": "object", "...": "..." \}, \\
\ \ \ \{"name": "green stems and leaves", "type": "object", "...": "..." \} \\
\ \ ], \\
\ \ "relationships": [ \\
\ \ \ \ "flowers growing in garden", \\
\ \ \ \ "flowers near building" \\
\ \ ] \\
\}%
}
\end{quote}

\subsection{Six-point grading system Criteria}

\begin{table}[H]
\centering
\caption{Six-point grading system and evaluation prompts}
\label{table-scoring}
\small 
\setlength{\tabcolsep}{8pt} 
\renewcommand{\arraystretch}{0.95} 
\begin{tabular}{@{}lp{9.2cm}@{}} 
\toprule
\textbf{Evaluation Dimensions} & \textbf{Key Elements of Prompts} \\
\midrule
Text Quality & Evaluate the clarity, grammatical accuracy, and relevance of the text. Check for duplications or irrelevant content. \\
Image Relevance & Assess whether visual elements precisely correspond to textual descriptions, rejecting generic/decorative images. \\
Cross-modal Consistency & Verify seamless integration between text and images, with coherent contextual transitions. \\
Task Completion & Measure the completeness of required actions in project-based tasks (e.g., all steps in tutorials). \\
Innovation & Evaluate originality in narrative approaches and visual storytelling techniques. \\
Harmful Content & Deduct 1 point for violent/offensive material (penalty criterion only). \\
\bottomrule
\end{tabular}
\end{table}

\end{document}